\documentclass[3p,final,11pt,authoryear]{elsarticle}

\usepackage{rotating}
\usepackage{amsmath}
\usepackage{tikz}
\usepackage{booktabs}
\usepackage{graphicx}
\usepackage{stackengine}
\usepackage{stackengine}
\usepackage{caption}
\usepackage{subcaption} 
\usepackage{amssymb}
\usepackage{amsthm}
\usepackage{tablefootnote}
\usepackage{algorithm}
\usepackage[noend]{algpseudocode}

\usepackage{multirow}
\usepackage{txfonts}
\usepackage{mathdots}

\newsavebox{\measurebox}

\usepackage{lineno}
\usepackage{array,calc}
\usepackage{lscape}
\usepackage{float}
\usepackage[unicode]{hyperref}
\hypersetup{pdfauthor={Name}}
\usepackage[nameinlink,capitalize]{cleveref}

\floatstyle{plaintop}
\restylefloat{table}

\usepackage{xcolor}
\usepackage{lscape}
\usepackage{setspace}
\usepackage[symbol,multiple]{footmisc}
\usepackage{array,calc}
\usepackage{makecell}

\usepackage{pdfrender}
\usepackage{amssymb}
    
\journal{Transportation Research Part E}

\begin{document}
\begin{frontmatter}
\title{Integrated Optimization of Automated Warehouse Operations and Last-Mile Transport for Differentiated On-Demand Delivery}
\author[1]{Xiaozhu Sun\footnote{Corresponding author}}
\author[1]{Bilal Farooq}

\address[1]{Laboratory of Innovations in Transportation (LiTrans),
Toronto Metropolitan University, Toronto, Canada}

\begin{abstract}
In the context of differentiated on-demand goods delivery services, this study proposes an integrated optimization method for automated guided vehicles (AGVs) based smart warehouse operations and the last-mile multi-modal transport. A deep reinforcement learning algorithm for multi-objective joint scheduling is designed to establish a dynamic connection between two systems, solving key challenges such as achieving high-throughput continuous order scheduling, meeting competing requirements, and improving the overall system sensitivity and adaptability. For warehouse optimization within this framework, we propose an improved algorithm based on multi-objective, Multi-Reward Machines--A* Guided Deep Q-Network (MORM-AGDQN), which combines service level, system cost, and external transportation demand. For external optimization, we propose an improved algorithm based on a Multi-Reward, Multi Head attention--Heterogeneous Capacity Vehicle Routing Problem (MRMH-HCVRP) framework, which incorporates the optimized scheduling order sequence and grouping, combined with customer location, demand, and priority, vehicle capacity, speed, and service range. The results show that the proposed framework significantly outperforms traditional methods, achieving 100\% on-time delivery rate for warehousing operations. After joint optimization, the average delivery time for the last mile was reduced by 29.3\% to 53.2\%, the total transportation distance was reduced by 46.4\%, the high-priority service rate was increased to over 92\%, and a balance was maintained between operating costs and customer satisfaction. 

\end{abstract}

\end{frontmatter}
\section{Introduction}
\label{intro}
Driven by the rapid growth of e-commerce and increasing consumer expectations for same-day or even instant delivery, urban logistics systems are undergoing a paradigm shift toward automation and intelligence \citep{HubnerAlexanderHermann2016Lmfa}. 
These technologies support dynamic storage, real-time order picking, and energy-efficient material handling, thereby enhancing both flexibility and responsiveness in intralogistics operations. More than 75\% of AGV applications are concentrated in indoor warehousing environments, driven by the increasing demand for e-commerce fulfillment and express delivery services \citep{YuanZhe2017BDfR}. To meet diverse market requirements including various order quantity, service priorities, and strict delivery deadlines, improving operation and transportation efficiency and reducing cost have become central objectives in warehouse design \citep{xu2022deep}.  Consequently, numerous studies have explored AGV scheduling, task assignment, and path planning algorithms with objectives such as time efficiency, energy consumption, or collision avoidance \citep{LiKunpeng2024Arlh,LiWenhao2022MPFw,ChenZhe2021ITAa}. Common AGV-based systems include Robotic Mobile Fulfillment Systems (RMFS) \citep{YangXiying2021Jooo}, Automated Storage and Retrieval Systems (AS/RS) \citep{BrezovnikSimon2015OoaA}, and intelligent/smart warehouses \citep{HonglinZhang2023Coot}.

Although within-warehouse logistics and last-mile transport delivery are deeply interconnected, they are traditionally optimized independently, leading to suboptimal global performance under dynamic constraints \citep{MasoudSherif2017AtCI}. While recent research emphasizes joint optimization \citep{YangXiying2025Jooo,WuJingwen2025Jooo,WangLi2023Joop}, traditional mathematical approaches struggle with the scalability required for stochastic demand and heterogeneous fleets. To address this gap, we propose a novel, integrated deep reinforcement learning (DRL) framework that coordinates automated warehouse operations with last-mile crowdsourced delivery. By using external environmental data to guide internal AGV scheduling and optimizing combined fleet routing, this approach balances multi-objective trade-offs to minimize costs and energy use while maximizing system autonomy and service levels. The main contributions of this work are as follows:

\begin{enumerate}

\item \emph{A bidirectional coordination framework} that synchronizes inner-warehouse AGV scheduling with heterogeneous last-mile fleet routing.
\item \emph{A unified multi-objective model} balancing complex trade-offs, including safety, energy, traffic, and customer priority—across the entire delivery chain.
\item \emph{A dual-network deep reinforcement learning architecture} combining CNNs, multi-head attention, and reward machines to jointly optimize dynamic logistics operations.

\end{enumerate}


The reminder structure of this paper is as follows: Section \ref{S:tBack} shows the related works in literature of review. Section \ref{ProbStat} presents the problem statement. Section \ref{Mechanism} establishes a new methodology that combines inner and external optimization mechanisms and shows mathematical functions and key constraints. 

Section \ref{casestudy} introduces the case study and simulations for this research. Section \ref{results} presents the results based on the process results, algorithm comparison, and key performance indicators (KPIs). Finally, Section \ref{conclusions} concludes the study and discusses future work.  

\section{Related work}
\label{S:tBack}
We introduce recent literature focusing mainly on AGV task allocation and path optimization in warehouses, order connection and path optimization for vehicles outside warehouses, and methods and applications of integrated optimization.


\subsection{Application of RL in AGV scheduling and path optimization in warehouse}\label{LR} 

In warehouse environments, AGV optimization relies on autonomous vehicles that iteratively refine control strategies through environmental interactions and feedback. Regardless of the algorithmic paradigm, defining appropriate objective functions is fundamental to overall performance. \cite{LiKunpeng2024Arlh} formulated a mixed-integer linear programming (MILP) model with valid inequalities to minimize total completion time, extendable to task priority, picking sequences, and congestion constraints. ~\cite{AndersenPer-Arne2020Tsri} employed a model-based policy optimization (MBPO) algorithm, while ~\cite{LiWenhao2022MPFw} proposed a prioritized communication multi-agent reinforcement learning (PICO-MARL) framework to enhance safety and reduce collision risk. Similarly, ~\cite{AgrawalAakriti2023RAAI} introduced a deep multi-agent reinforcement learning model (RTAW) to minimize travel delays, and ~\cite{ChenZhe2021ITAa} presented a marginal-cost metaheuristic for multi-capacity carriers to reduce delivery costs. ~\cite{KrnjaicAleksandar2022SMRL} combined AGVs and human operators via shared actor–critic networks to improve task efficiency, though travel distance and energy use were not explicitly addressed. Furthermore, ~\cite{ChengBayi2024Drld} optimized batch order scheduling (EDRL-OBOS) using deep reinforcement learning (DRL) to lower operational costs; however, worker efficiency remained unevaluated. 

In double-objective optimization, most studies emphasize internal warehouse indicators. ~\cite{WuShao-Ci2024DRLf} proposed a DQN-based regret and marginal-cost task assignment (RMCA) method to minimize cost and maximize allocation efficiency. ~\cite{DouJiajia2015GSaR} integrated reinforcement learning with genetic algorithms to jointly minimize distance and travel time, while ~\cite{LiMaojiaP.2019TSbA} developed a Rainbow DQN dispatching system to reduce mean task time and collisions. ~\cite{LeeHyeokSoo2021MRPO} compared Q-learning and dynamic-Q (Dyna-Q) for path efficiency, and ~\cite{YangYang2020Mppb} improved DQN convergence through empirical playback. However, traditional A* algorithms still perform poorly in dynamic obstacle scenarios, thereby making adaptive local path planning a persistent challenge.

Recent studies have extended DRL frameworks to multi-objective optimization for dynamic and uncertain warehouse environments. ~\cite{KamoshidaRyota2017AoAG} optimized travel distance, operation time, and congestion avoidance using high-dimensional map information; however, this was limited to static layouts. ~\cite{HaWonYong2021AWSU} explored the trade-offs among safety, time, and efficiency through GA-based collision indices, whereas ~\cite{SartorettiGuillaume2019PPvR} scaled AGV coordination from four to 1024 units by combining reinforcement and imitation learning. ~\cite{LiShuo2022RoOA} employed proximal policy optimization (PPO) with refined density, distance, and step-related rewards to balance safety and success rate; however, this was constrained by small map scales and limited fleets. Only a few works have rigorously applied multi-objective reinforcement learning (MORL) to warehouse path planning. Among them, ~\cite{YangJintao2023RoIW} proposed a multi-policy Q-learning (MPQRL) framework that balanced multiple objectives, achieving higher coverage ratios and lower average expected utility in AGV routing.

Most existing studies target transport carriers that move entire shelves, which is suitable for flexible manufacturing or smart warehouse settings. Medium-sized warehouses typically involve fewer than 30 vehicles, whereas large-scale or simulated environments can test hundreds. However, these works largely focus on single-load AGVs. Only Chen et al. ~\cite{ChenZhe2021ITAa} examined multi-load scenarios involving multiple item boxes with identical specifications. Multi-load optimization fundamentally differs from single-load problems, combining traveling salesman and dynamic optimization challenges. The position, status, and timing of each order influence AGV decisions, and system updates cause dynamic shifts in reinforcement learning strategies. Despite progress, prior studies have not clearly defined multi-objective formulations, often limiting the analysis to performance indicators rather than explicit objective trade-offs. 

\subsection{Application of RL in vehicle routing optimization external the warehouse}\label{LR}

Efficient vehicle routing beyond the warehouse boundary is a critical component of end-to-end logistics optimization. Although warehouse automation has advanced through AGV-based systems, external transportation remains a major determinant of delivery cost, emissions, and customer satisfaction. The growing complexity of urban logistics, driven by real-time traffic fluctuations, uncertain order arrivals, and heterogeneous fleet compositions, requires adaptive mechanisms capable of learning from dynamic environments. In this context, deep reinforcement learning (DRL) has emerged as a promising paradigm that enables agents to iteratively refine routing policies based on environmental feedback.

Unlike traditional heuristic or mixed-integer approaches, DRL-based routing frameworks learn adaptively from traffic conditions, vehicle energy status, and stochastic demand. The integration of graph-attention structures (GAs) with reinforcement learning has proven effective for capturing spatial dependencies and improving routing efficiency; however, scalability challenges remain \citep{ZhangKe2023Garl}. ~\cite{WangChao2025ADRL} developed an attention-augmented DRL mechanism for electric vehicle routing, enhancing energy management and charging efficiency under dynamic grid conditions. Likewise, ~\cite{LiJingwen2022HAfS} proposed a heterogeneous attention network for complex pickup-and-delivery scenarios, effectively modeling spatial–temporal dependencies. These studies highlight the superior adaptability and robustness of DRL-based models over classical optimization; however, issues such as hyperparameter sensitivity and training scalability persist.

Recent advances have refined DRL for routing under capacity and uncertainty constraints. ~\cite{bai2025improved} proposed an improved DRL model for capacity-constrained VRP, achieving higher route efficiency and solution quality on benchmark datasets but remaining parameter-sensitive. Similarly, ~\cite{KadyrovShirali2025Drlf} introduced a DRL framework for dynamic VRP incorporating demand and traffic uncertainty. Their attention-based encoder adapts routing strategies in real time, minimizing response time and operational cost at the expense of higher training complexity.

In contrast to general VRP research, an emerging stream focuses on last-mile delivery optimization, particularly in urban and crowdsourced logistics. ~\cite{ElAmraniAmineMohamed2025ADRL} developed a traffic-aware DRL framework integrating public transport systems in Casablanca, reducing delays and energy use but facing limited scalability. In dynamic crowdsourcing contexts, several studies \citep{SalehZead2025CRLA, SilvaMarco2022DRLf, XiangChuankai2024CDRL, daSilvaMarcoAurelioCosta2023Drlf, VeraJoseManuel2019DRLf} proposed DRL-based coordination mechanisms for fluctuating demand, variable courier availability, and uncertain capacities. For instance, ~\cite{XiangChuankai2024CDRL} presented a centralized attention-based multi-vehicle coordination network that optimized task allocation and routing under stochastic courier arrivals. While~\cite{daSilvaMarcoAurelioCosta2023Drlf} addressed stochastic dynamic demand,  ~\cite{VeraJoseManuel2019DRLf} incorporated probabilistic order arrivals to better model uncertainty. ~\cite{DengJianjun2024MMES} further formulated a multi-objective DRL framework for green last-mile delivery, balancing cost, emissions, and timeliness.

Overall, the applications of DRL in vehicle routing illustrate a shift toward data-driven, adaptive, and sustainable logistics optimization. However, challenges remain in multi-objective convergence, balancing exploration and exploitation under stochastic conditions and coordinating heterogeneous fleets. These limitations motivate the need for integrated frameworks that jointly optimize in-warehouse AGV scheduling and out-of-warehouse delivery routing to advance a unified and intelligent logistics ecosystem. 

\subsection{Integrated Process Optimization}\label{MORL op}

With the growing demand for same-day and on-demand delivery, the integration of smart warehousing with intelligent last-mile distribution has become a pivotal direction for improving overall logistics performance. Recent studies have shown that jointly optimizing in-warehouse and external delivery processes yields superior results compared with independent optimizations. This integration enhances the coordination between order picking, replenishment, routing, and delivery scheduling, thereby improving throughput, delivery timeliness, and resource utilization. 

\cite{YangXiying2025Jooo} proposed a joint order allocation and shelf scheduling strategy for Kiva Systems and RMFS, employing mixed-integer programming (MIP) and annealing heuristics to minimize completion time and increase throughput. \cite{WuJingwen2025Jooo} optimized RMFS joint picking and replenishment through a five-stage MIP model, substantially reducing operating costs. However, both studies were limited to internal processes and did not extend to last-mile transportation optimization. \cite{JiangZhong-Zhong2024Jooo} developed an ergonomically optimized system integrating order picking and delivery, enhancing coordination, sustainability, and worker well-being, although the framework assumed deterministic human behavior and lacked scalability. 

Expanding toward end-to-end coordination, \cite{WangLi2023Joop} established a joint scheduling and routing framework that connects warehouse operations with crowdsourced last-mile delivery. Their integrated model improved on-time delivery rates and reduced transport time under congestion through dynamic coordination. Similarly, \cite{SchubertDaniel2018Iopa} and \cite{KUHN20211003} emphasized the co-optimization of order batching, picking, and routing using an iterative local search (ILS) metaheuristic, improving same-day delivery performance. \cite{RajG.2024Smoi} incorporated stochastic demand factors into the joint optimization of order picking and vehicle routing using MIP, providing more robust scheduling under uncertainty. \cite{GuQiuchen2024Isoo} proposed an integrated scheduling framework that combines order picking, batching, and courier assignment to minimize total fulfillment time and improve responsiveness.

Collectively, these studies demonstrate the advantages of integrated optimization frameworks in synchronizing warehouse operations and delivery processes. Joint decision-making can mitigate congestion effects, enhance temporal coordination, and improve resource efficiency across the supply chain. Nonetheless, most current approaches rely on deterministic or heuristic optimization, with limited application of adaptive reinforcement learning or stochastic decision-making, leaving significant potential for dynamic, data-driven integration in real-time logistics systems. 

\section{ Problem Statement}\label{ProbStat} 

\begin{figure}[!ht]
\centering
  \includegraphics[width=1.0\textwidth]{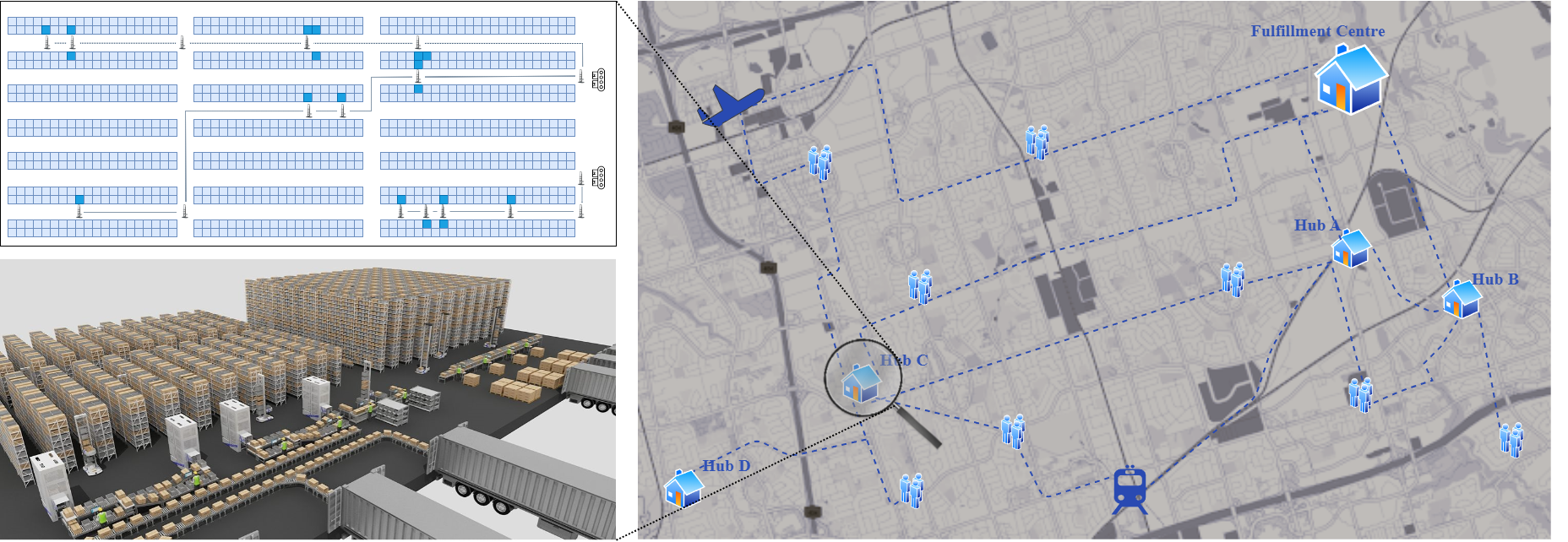}
  \caption{Integration of warehouse operations and last-mile delivery}\label{fig:map}
\end{figure}

This study investigates the integrated goods delivery problem primarily applied to systems such as smart warehouse services for e-commerce delivery, express delivery distribution centres, just-in-time (JIT) delivery, and urban warehouse to client's door delivery services. In these scenarios, the configuration of the warehousing centre is simpler than that of the fulfilment centre, mostly consisting of pre-packaged goods or goods requiring only simple re-packaging. Based on the optimization objectives of integrated systems, we physically divided the main process into two parts: the internal components and workflow of the smart warehouse and the external vehicle routing and delivery, as shown in Figure \ref{fig:map}.

\subsection{Internal Components and Workflow}

\begin{figure}[!ht]
\centering
  \includegraphics[width=1.0\textwidth]{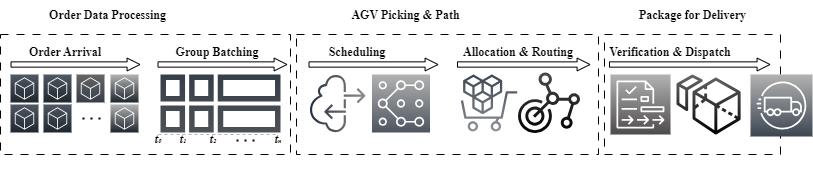}
  \caption{Internal Warehouse Workflow}\label{workflow}
\end{figure}

To achieve high automation and efficiency, an intelligent warehouse system relies on integrated infrastructure, including multi-tier racks, multi-capacity automated guided vehicles (AGVs), workstations, and conveyors. The warehouse layout features optimized single- or double-lane aisles where goods are stored in specific rack grid cells by SKU \citep{ZhangZheng2023AoAG}. Coordinated wirelessly by a central warehouse management system (WMS), AGVs autonomously retrieve and transport items to workstations while maintaining optimal energy levels. Finally, human or robotic operators verify and package the goods before conveyor belts dispatch them to the outbound area.
The operational workflow of an intelligent warehouse follows a systematic and data-driven process, as illustrated in Figure \ref{workflow}, and can be summarized as follows:

\begin{enumerate}
\item Order Reception and Data Processing: The system continuously receives real-time customer orders and associated metadata. To optimize processing efficiency, incoming orders are grouped into batches within predefined time windows.
\item Dynamic Order Scheduling and Prioritization: The system analyzes each batch using proposed optimization algorithms to determine the optimal picking sequence and expected completion time.
\item Task Allocation and Route Planning: Based on the scheduling results, tasks are assigned to available AGVs. Each AGV path is then optimized using the proposed mechanism.
\item Automated Picking and Transportation: The AGVs execute their assigned missions by retrieving multiple items, enabled by their multi-capacity handling mechanism, and delivering them to the designated pick-up station or workstation.
\item Verification, Packaging, and Dispatch: At the workstation, each order undergoes verification against digital records. Items are repackaged if required and then sequentially placed onto the conveyor system for dispatch toward the warehouse exit or delivery area.
\end{enumerate}

\subsection{External Vehicle Routing and Delivery}

\begin{figure}[!ht]
\centering

\begin{subfigure}{0.45\textwidth}
  \centering
  \includegraphics[height=5cm]{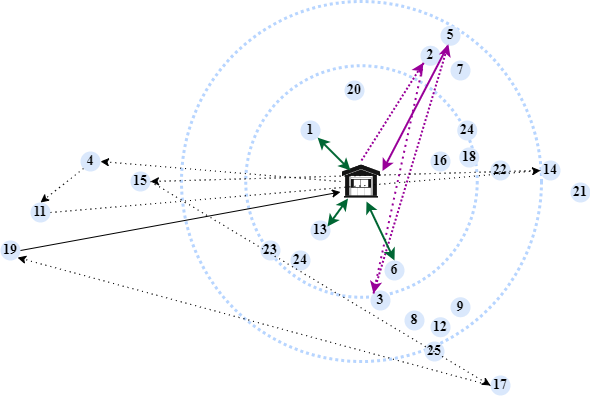}
  \caption{Before}
  \label{fig:sub1}
\end{subfigure}%
\hfill
\begin{subfigure}{0.45\textwidth}
  \centering
  \includegraphics[height=5cm]{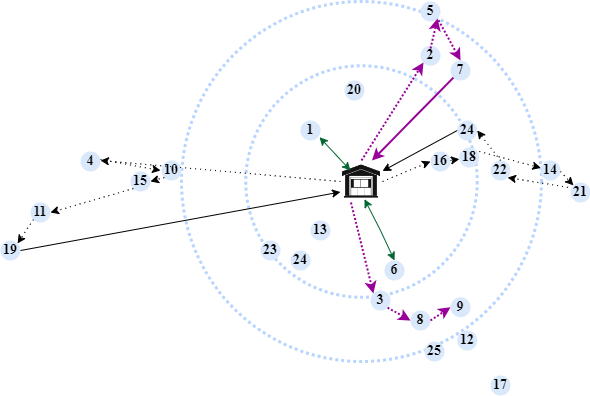}
  \caption{After}
  \label{fig:sub2}
\end{subfigure}

\caption{External Vehicle Routing Differences}
\label{externalVRP}

\end{figure}

Timely last-mile delivery is crucial for high-priority and geographically dispersed orders, requiring precise coordination with internal warehouse operations to achieve JIT performance. As this tightly coupled system is influenced by dynamic factors like traffic, capacity constraints, and time windows, we frame it as a multi-objective optimization problem for heterogeneous fleets \citep{LiJingwen2022HAfS,SarangiSubrat2023Hmpa,PericNikica2025Oohl}. As illustrated in Figure \ref{externalVRP}, the system matches vehicles, which vary in speed, capacity, and mileage, to delivery clusters to balance timeliness, cost, and resource utilization. While optimizing mixed-fleet routing is a computationally complex, NP-hard knapsack problem \citep{CacchianiValentina2022Kp—A}, we simplify our scope to ensure feasibility. Rather than incorporating complex variables like EV charging or vehicle leasing, we group fleets by service distance, capacity, and speed \citep{MengFanchao2019Mmko}, deferring fully hybrid fleet scenarios to future research.

\subsection{Key Notations and Definition}
The following Tables \ref{tab:internal_notations} and \ref{tab:external_notations} show all key notations and definitions for integrated optimization of warehouse and last-mile delivery.

\begin{table}[h]
\footnotesize

\caption{Key Notations for Internal AGV Optimization}
\centering
\begin{tabular}{p{0.15\textwidth}p{0.15\textwidth}p{0.5\textwidth}}
\hline
\textbf{Type} & \textbf{Symbol} & \textbf{Description} \\

\hline
 & $\mathcal{K}$ & Set of AGVs \\
 & $\mathcal{V}$ & Set of all nodes (storage racks, pick stations) \\
Set  & $\mathcal{O}$ & Set of orders\\
 & $k$ & Index for AGVs, $k \in \mathcal{K}$ \\
 & $i, j$ & Indices for nodes, $i, j \in \mathcal{V}$ \\
\midrule
 & $x_{ij}^k$ & 1 if AGV $k$ travels from node $i$ to $j$; 0 otherwise \\
Decision & $p_{j}^k$ & 1 if AGV $k$ picks order at node $j$; 0 otherwise \\
 Variable& $x_{i}^k$ & 1 if AGV $k$ visits node $i$; 0 otherwise \\
  & $h_i$ & 1 if no collision at node $i$; 0 if collision occurs \\
 
\hline
 & $Q_k$ & AGV net weight $k$ \\
 & $W_k$ & Weight limitation of AGV $k$ \\
 & $E_k$ & Battery or energy limit of AGV $k$ \\
& $e_{ij}^k$ & Energy consumed by AGV $k$ from node $i$ to $j$\\
 & $d_{ij}^k$ & Distance from node $i$ to $j$ for AGV $k$ \\
Parameter & $m_k$ & Capacity (maximum number of orders) for AGV $k$ \\
& $w_{ij}^k$ & Load weight carried by AGV $k$ from node $i$ to $j$\\

 & $c^{ABC}_j$ & Inventory cost for an ABC-class order at node $j$ \\
 & $d^{out}_v$ & Estimated last-mile distance from warehouse to customer $v$ \\
 & $\gamma_{\mathrm{lan}(t)}$ & LAN traffic congestion multiplier at time $t$ \\
\hline
\end{tabular}
\label{tab:internal_notations}
\end{table}

\begin{table}[h]
\centering
\caption{Key Notations for External Vehicle Routing}
\footnotesize
\begin{tabular}{p{0.1\textwidth}p{0.1\textwidth}p{0.6\textwidth}}
\hline
\textbf{Type} & \textbf{Symbol} & \textbf{Description} \\
\hline
 & $\mathcal{H}$ & Set of all vehicles \\
 & $\mathcal{C}$ & Set of all customers \\
Set & $\mathcal{N} = \{0\} \cup \mathcal{C}$ & Set of all nodes (0 is the depot/warehouse) \\
 & $h$ & Index for vehicles, $h \in \mathcal{H}$ \\
 & $u, v$ & Indices for nodes (customers/depot), $u, v \in \mathcal{N}$ \\
 \hline
Decision & $y_{uv}^h$ & 1 if vehicle $h$ travels from node $u$ to $v$, 0 otherwise \\
 Variable & $\delta_h$ & 1 if vehicle $h$ is used, 0 otherwise \\
 & $\mathrm{served}_u$ & 1 if the customer is served, 0 otherwise\\
\hline
 & $t$ & Time variable \\
 & $w_u^{\mathrm{pr}}$ & Priority weight for customer $u$ \\
 & $T_u^{\mathrm{dead}}$ & Deadline for serving customer $u$ \\
 & $T^{pro}_u$ & Processing time at customer $u$ \\
&$T^{ar}_u$ & Arrival time at customer $u$ \\
&$T^{ser}_u$ & Service time at customer $u$ \\
& $T_u^{\mathrm{ideal}}$ & ideal service time for customer $u$\\
Parameter & $T_{0u}^{h} (t)$ & Total travel time from depot $0$ to customer $u$ by vehicle $h$ departing at time $t$ \\
&$T_{uv}^h(t)$ & Travel time for vehicle $h$ from depot $u$ to $v$ starting at time $t$ \\
& $D_{uv}$ & Distance between node $u$ and $v$ \\
 & $f_h^{base}$ & Base speed of vehicle $h$ \\
 & $\beta_h(t)$ & Time-period impact factor (peak hours) for vehicle $h$ \\
 & $\gamma_h(r)$ & Regional impact factor (e.g., downtown) for vehicle $h$ in region $r$ \\
 & $\lambda_{hrt}$ & Congestion propagation factor for vehicle $h$ in region $r$ at time $t$ \\
 & $\lambda_o $& Opportunity cost weighting coefficient \\
 & $z_h$ & Cost per unit distance for vehicle $h$ \\
 & $F_h$ & Fixed cost for using vehicle $h$ \\
 & $D_h^{\max}$ & Maximum travel distance for vehicle $h$ \\
 & $Q_h$ & Max Capacity of vehicle $h$ \\
 & $Q_h^{used}$ & Actual load used of vehilce $h$ \\
 & $q_v$ & Demand of customer $v$ \\

\hline
\end{tabular}
\label{tab:external_notations}
\end{table}

\section{ Methodology }\label{Mechanism} 

Unlike traditional solvers such as Gurobi or Google OR-Tools, which rely on simplified hierarchical optimization and struggle with large-scale, dynamic Pareto front exploration \citep{SAGHAND2022105549}, Multi-Objective Reinforcement Learning (MORL) uses vectorized rewards to manage conflicting goals. To leverage this, our study proposes a unified MORL framework that integrates internal AGV scheduling (balancing energy, safety, and inventory) with external fleet routing (balancing cost, time, and service quality). As these layers interact dynamically (e.g., prioritizing urgent orders may increase transport costs), our framework utilizes a dynamic reward mapping function to translate static constraints and real-time conditions into multi-objective feedback. Through neural networks and attention heads, agents adaptively learn these trade-offs to maximize cumulative rewards and maintain system-wide operational balance.

\subsection{ Multi-objective Optimization Function }\label{optimization define} 
For a static system, a multi-objective optimization function can be generally described as:
\begin{equation}
        \min  F_s(x) =  (f_1(x),f_2(x),...,f_M(x))
 \label{utilityMLM}
\end{equation}
where $ F_s(x)$ is the multi-objective function, $x$ denotes the static variable,$f_1(x),f_2(x),...,f_M(x)$ are all the single-objective functions for optimization goals, and $x = (x_1, x_2,..., x_n) $ is the decision vector. 

When the definition of the problem is dynamic and factors change over time, a dynamic multi-objective optimization based function can be improved as follows:

\begin{equation}
        \min  F_d(x,t) =  (f_1(x,t),f_2(x,t),...,f_N(x,t))
 \label{utilityMLM}
\end{equation}
where $F_d(x,t)$ is the dynamic multi-objective function with time,$f_1(x,t),f_2(x,t),...,f_N(x,t)$ mean all the single objective functions for optimization goals, $x = (x_1, x_2,..., x_n) $ is the decision vector and $t$ is the
time window of the system update, and the general approach is to evenly divide the running time into n time steps. 
In this study, the mixed objective function is used to sum up these two situations into a state which performs a summation in each time window and then updates the current optimal results into the next time window. This approach preserves the temporal granularity of short-term decisions while maintaining global consistency across the full scheduling horizon, offering a balanced trade-off between responsiveness and overall optimality. For each time zone, the objective function is:

\begin{equation}
        \min  F(x,t) = \min (F_s(x) + F_d(x,t)) 
 \label{utilityMLM}
\end{equation}

\subsubsection{Multi-Objectives for Internal Warehouse System}\label{internal MO} 
The evaluation of AGV performance in intelligent warehousing and transportation systems primarily focuses on effective cost control, order service quality, and safe and stable operations. These objectives are inherently interdependent, and trade-offs may occur during the optimization process. Furthermore, each objective function is influenced by both static and dynamic factors, including the type and operating status of the AGV, warehouse layout and environmental conditions, and external influencing factors (local traffic conditions within the logistics network and changes in order structure or scheduling priorities). These factors collectively determine the efficiency, reliability, and adaptability of internal scheduling and path planning.

\paragraph{a. Energy Consumption of AGVs}
To minimize integrated costs, reducing AGV energy consumption is a fundamental sub-objective of our model. As dynamic energy expenditure is driven by vehicle velocity, distance, and payload weight, we formulated the energy consumption objective as a linear function of transportation distance and total weight (including both intrinsic vehicle mass and dynamic load). This formulation aligns with established research demonstrating a proportional relationship between distance, load, and energy use in automated material handling systems \citep{LiangQiu2015HArp}:
\begin{equation}
        f_{En}(x,t) =  (f_d(x,t),f_g(x,t),f_v(x,t))
 \label{utilityMLM}
\end{equation}
where $f_{En}(x,t)$ represents the AGV energy consumption function, and $f_d(x,t),f_g(x,t),f_v(x,t)$ define the distance, weight, and velocity-related functions by time, respectively. Because capturing the instantaneous speed of AGVs or vehicles in detail has little impact on the overall process in the integrated external optimization, this study assumes that the transportation vehicles operate at an average speed. We expand the energy-related objective function and constraints as follows: 

\begin{equation}
\min f_{En} = 
\sum_{k=1}^{K} \sum_{i=0}^{n} \sum_{j=0}^{n} 
d_{ij}^k\, x_{ij}^k\, f(Q_k + w_{ij}^k)
\label{eq:energy-objective}
\end{equation}

\text{s.t. } \quad
\begin{equation}
w_{ij}^k \le Q_k\, x_{ij}^k,
\qquad 
\forall i,j \in V,\; k = 1,\ldots,K
\label{eq:load-constraint}
\end{equation}

\begin{equation}
\sum_{i=0}^{n} x_{i_{O}D} 
= 
\sum_{i=0}^{n} x_{i_{D}O}
\le K,
\qquad
\label{eq:routing-constraint}
\end{equation}

\begin{equation}
x_{ij}^k \in \{0,1\}, \quad \forall i, j \in V,\; k = 1, 2, \ldots, K
\label{eq:one route}
\end{equation}

\begin{equation}
\sum_{j=1}^{n} \sum_{i=0}^{n} x_{ij}^k \le m_k,
\qquad \forall k=1,\ldots,K.
\label{eq:capacity}
\end{equation}

\begin{equation}
\sum_{i=0}^{n} \sum_{j=0}^{n} e_{ij}^k \cdot x_{ij}^k \leq E_k,
\qquad \forall k = 1,\ldots,K
\label{eq:battery-constraint}
\end{equation}

where \ref{eq:load-constraint} states that AGVs can pick up multiple goods over a distance, but the total weight of the goods cannot exceed the weight limit. Constraint \ref{eq:routing-constraint} indicates that a maximum of K AGVs can be used, and each AGV should start and end from the initial point and the fixed position in front of the conveyor belt. The binary variable \ref{eq:one route} represents the decision variable, that is, whether the path between two nodes is executed or not. Constraint \ref{eq:capacity} indicates the limit on the number of items that the AGV can carry. The ${i,j\in V}$ are the origin and destination nodes in the warehouse, and each node j corresponds to a pickup location, delivery location, or depot. Finally, constraint \ref{eq:battery-constraint} maintains the battery usage for AGVs.

\paragraph{b. Inventory Cost for Orders}\label{Inventory} 
In this study, inventory is classified into A, B, and C categories based on operational importance and financial value (e.g., turnover rate, carrying cost) using ABC analysis principles \citep{RavinderHandanhalV2016AAFI, TeunterRuudH.2010ACSL, ValdiviaSeminarioCarlos2023AoAC}. Before AGV pickup, waiting orders incur time-sensitive backlog costs determined by their waiting duration, inventory class, and commodity value. By integrating this composite cost function into the multi-objective reinforcement learning (MORL) framework, the model optimizes AGV scheduling to minimize storage and waiting costs while ensuring timely delivery. The specific definitions, expansions, and constraints are as follows:

\begin{equation}
        f_{In}(x,t) =  (f_{abc}(x),f_{st}(x,t))
 \label{utilityMLM}
\end{equation}
\begin{equation}
f_{abc}(x)=
\begin{cases}
f_{a}(x) &  \text{A class of inventory } \\
f_{b}(x) &  \text{B class of inventory }  \\
f_{c}(x) &  \text{C class of inventory } \\
\end{cases}
 \label{14}
\end{equation}
where $ f_{In}(x,t)$ represents the inventory cost due to order waiting, $f_{abc}(x)$ is a three-segment function depending on product classification of $f_{a}(x),f_{b}(x),f_{c}(x)$ and $f_{st}(x,t)$ defines an inventory cost general function by time.

\begin{equation}
\begin{aligned}
\min f_{In} &= 
    \sum_{k=1}^{K} \sum_{j=1}^{n} c^{ABC}_{j}\, p_{j}^k 
\end{aligned}
\label{eq:inventory-objective}
\end{equation}

\text{s.t. } \quad
\begin{equation}
\sum_{i=0}^{n} x_{ij}^k = p_{j}^k,
\qquad \forall j = 1,\ldots,n,\; k = 1,\ldots,K.
\label{eq:linking}
\end{equation}

\begin{equation}
\sum_{j=1}^{n} p_{j}^k \le m_k,
\qquad \forall k = 1,\ldots,K.
\label{eq:layer-capacity}
\end{equation}

\begin{equation}
p_{j}^k \in \{0,1\}, \quad 
x_{ij}^k \in \{0,1\},
\qquad \forall i,j=0,\ldots,n,\; k=1,\ldots,K.
\label{eq:binary-vars}
\end{equation}

Equation \ref{eq:inventory-objective} defines the total inventory-related cost incurred by the system, weighted by the ABC classification of each order. Constraint \ref{eq:linking} enforces the linking between routing decisions and inventory assignments, ensuring that an order $j$ is considered picked up by AGV $k$ only when the vehicle physically visits the corresponding node. Constraint \ref{eq:layer-capacity} specifies the maximum number of goods that each AGV can carry simultaneously, reflecting the structural limitation of the vehicle composed of $m_k$ storage layers, where each layer can accommodate only one order. Constraint \ref{eq:binary-vars} defines the binary trait of both routing variables and pickup-assignment variables.

\paragraph{c. External Influence}\label{External} 
External factors outside the warehouse, including customer priority, destination distance, outbound transport mode, and the real-time status of the local transportation network (LAN), significantly impact the overall service efficiency of the integrated warehouse and last-mile delivery system. These factors are jointly represented in the external-influence objective function $f_{Ex}$, which aggregates subcomponents related to customer priorities, order destination distance, and LAN traffic conditions (e.g., traffic state and weather):

\begin{equation}
        f_{Ex}(x,t) =  (f_{de}(x,t),f_{od}(x),f_{lan}(x,t))
 \label{utilityMLM}
\end{equation}
where $f_{Ex}(x,t)$ represents the external multi-factor function and $f_{de}(x,t),f_{od}(x),f_{lan}(x,t)$ are delay function, order destination calculation, transport mode, and LAN status, respectively.\\

\textit{c.a. Delay by customer priority:} To improve the system service level and response rate, a function of delay cost changing over time is designed based on the priority classification of service objects \citep{cdiproquestreports1117364637}. Without the loss of generality, this study applies a four-level classification and its change curve to calculate the dynamic delay loss generated by the order after the deadline in proportion to the total delay: 
\begin{equation}
f_{de}(x,t)=
\begin{cases}
f_{A}(x,t) &  \text{A class of customers } \\
& \vdots \\
f_{D}(x,t) &  \text{D class of customers } \\
\end{cases}
 \label{eq:delay-class}
\end{equation}
where $ f_{de}(x,t)$ represents the delay cost due to the classification of customer ratio and time, including several classes such as $f_{A}(x),f_{B}(x),f_{C}(x),f_{D}(x)$ or more, depending on the sale system or delivery requirements.\\

\textit{c.b. Order destination:}
The outbound transportation distance depends on the customer's location relative to the warehouse. Longer distances result in increased travel times and transport costs:
\begin{equation}
f_{od}(x) = \sum_{v=1}^{n} d^{out}_{v}\, p_{v},
\label{eq:order-destination}
\end{equation}
where $d^{out}_{v}$ is the estimated last-mile distance from the warehouse to customer $v$, and $p_v$ is a pickup indicator.

\textit{c.c. LAN traffic condition:}
The dynamic local area network traffic conditions directly influence the travel duration. Congestion levels are categorized as jammed, high volume, medium volume, and low volume, each associated with a different travel time multiplier. We do not consider detailed traffic changes on the road but consider the average level to send notifications to the warehouse inner system:
\begin{equation}
f_{lan}(x,t) = \sum_{v=1}^{n} \gamma_{\mathrm{lan}(t)}\, d^{out}_{v}\, p_{v},
\label{eq:lan-traffic}
\end{equation}
where $\gamma_{\mathrm{lan}(t)}$ is the congestion multiplier at time $t$. The delay function in \ref{eq:delay-class} reflects different classes of priority penalties over time. Equation \ref{eq:order-destination} describes how outbound distance contributes to travel requirement, while constraint \ref{eq:lan-traffic} captures dynamic LAN congestion and its multiplier effect on travel time. 

\paragraph{d. Safety}\label{safe} 
The safety of the system is reflected in the fact that AGVs do not collide with obstacles or other vehicles during transportation and create deadlocks. To avoid congestion, in a time unit, when another AGV appears within a calibrated distance near an AGV, the carrier is alerted:
\begin{equation}
        f_{Sa}(x,t) =  (f_{h}(x),f_{con}(x,t))
 \label{utilityMLM}
\end{equation}
where $f_{Sa}(x,t)$ represents the safety insurance function avoiding hit$f_{h}(x)$ and congestion$f_{con}(x,t)$.
\subsubsection{Multi-Objectives for External Multimodal Delivery System}\label{external MO} 
The evaluation of heterogeneous fleet operation performance mainly considers factors such as reasonable fleet allocation, vehicle operating costs, timely order delivery, and the total time. Fleet allocation is influenced by customer group distribution, transportation costs primarily involve transportation distance and traffic conditions, and order delivery requires meeting priority service objectives, warehouse order picking orders, and time window constraints. Therefore, these objectives are interconnected during optimization process. To formally capture this, we define the objective vector $F_{VRP}(x,t)$ as in Eq.~\eqref{VPR}, where $x$ denotes the routing decision variable, $t$ represents time, $f_{Pr}(x,t)$ is the priority-related function, and $f_{Tr}(x,t)$ is the travel time function.  

\begin{equation}
        F_{VRP}(x,t) =  (f_{Pr}(x,t),f_{Tr}(x,t))
 \label{VPR}
\end{equation}

\paragraph{a. Priority Service}
The completion rate of tiered orders determines the quality of the customer service. Currently, many shipping services promise same-day or half-day delivery to meet the service needs of high-priority customers (e.g., paid expedited shipping, membership fees, and VIP customer benefits). The evaluation criterion is whether the goods can reach the customer's location within the deadline or be delivered as quickly as possible. The goal is to minimizes the transport time for higher-priority customers and prevent violating deadlines. The specific definition formula, expansion formula, and constraints are as follows:

\begin{equation}
        f_{Pr}(x,t) =  (f_{ar}(x) , f_{ser}(x))
 \label{utilityMLM}
\end{equation}

where $f_{ar}(x)$ proposes a priority-based function to ensure a quick arrival and $f_{ser}(x)$ ensures no delay.
\begin{equation}
\begin{aligned}
\min f_{Pr} &= 
    \sum_{u\in \mathcal{C}}^{n} w_u^{\mathrm{pr}} T^{ar}_u + P \sum_{u \in C} 1 \{ T_u^{\mathrm{ser}} > T_u^{\mathrm{dead}} \}
\end{aligned}
\label{eq:priority}
\end{equation}

\begin{equation}
T^{ar}_u = t_u + T_{0u}^{h} (t)
\label{eq:arrive}
\end{equation}

\text{s.t. } \quad
\begin{equation}
T^{ser}_u=T^{ar}_u +T^{pro}_u \le T_u^{\mathrm{dead}},
\qquad \forall u \in \mathcal{C}
\label{eq:deadline}
\end{equation}
The optimizer \ref{eq:priority} strongly prefers early arrival for the customer, and the arrival time for each customer is shown in \ref{eq:arrive}. $T_{ou}^{h} (t)$ is the total travel time from warehouse node $o$ to customer $u$ using vehicle $h$. Constraint \ref{eq:deadline} imposes a hard constraint on service time. 

\paragraph{b. Total Travel Time}
Total last-mile transport time is a crucial metric in vehicle routing that directly reflects our method's integrated effectiveness. With heterogeneous fleets, delivery times vary significantly based on vehicle constraints (speed, mileage, capacity) and real-time conditions. For instance, while faster vehicles excel in smooth traffic, alternative vehicles utilizing low-speed or bike lanes often perform better during congestion. To account for this variance, our transport time calculations incorporate each vehicle's base and effective speeds, alongside dynamic traffic factors such as peak hours, regional density, and congestion cascading effects \citep{FigliozziMiguel2010VRPf,HornMarkE.T.2006OVRa}.
The objective function and constraints are:

\begin{equation}
\min f_{Tr} = 
\sum_{h\in\mathcal{H}} \sum_{u=0}^{n} \sum_{v=0}^{n} 
T_{uv}^h(t_{uv}^h) y_{uv}^h
\label{eq:energy-objective}
\end{equation}

\text{s.t. } \quad
\begin{equation}
\sum_{h=1}^m \sum_{v=0}^n y_{uv}^h  = 1 \quad \forall u \in \mathcal{C} \\
\label{each node}
\end{equation}
\begin{equation}
\sum_{v=0}^n y_{uv}^k  = \sum_{v=0}^n y_{vu}^h \quad \forall h, \forall u
\label{vehicle flow}
\end{equation}
where $f_{Tr}$ calculates the total travel time for all vehicles and customers. Constraint \ref{each node} ensures that each customer is visited exactly once, and constraint \ref{vehicle flow} follows the flow conservation for each vehicle. Travel time for vehicle $h$ on arc $(u,v)$ departing at time $t$ is:

\begin{equation}
T_{uv}^h(t) = \frac{D_{uv}}{f_h^{eff}(t, r(v))}
\label{eq:travel-time}
\end{equation}

where $r(v)$ is the region of node $v$. The effective speed of vehicle $h$ at time $t$ in region $r$ is calculated as:

\begin{equation}
f_h^{eff}(t,r) = f_h^{base} \cdot \beta_h(t) \cdot \gamma_h(r) \cdot (1 - \lambda_{hrt})
\label{eq:effective-speed}
\end{equation}

where $f_v^{base}$ is the base speed of vehicle $h$, $\beta_h(t)$ shows time period impact factor (speed reduction during peak hours), $\gamma_h(r)$ represents regional impact factor (slower speed in downtown areas), and $\lambda_{hrt}$ is the congestion propagation factor (cascading traffic effects).

\subsection{Dual-Architecture  Mechanism}\label{optimization method} 
To integrate AGV scheduling and path planning within automated warehouses and the Heterogeneous Capacity Vehicle Routing Problem (HCVRP) outside warehouses into a dynamic decision-making process, this study models the problem in both spaces as a Dual-Architecture Markov Decision Process (MDP). It relies on effective reinforcement learning tools to establish connections between the two decision processes, allowing the information flow and influencing factors to function effectively in both systems.

\subsubsection{RL Architecture for Internal Warehouse System}
The scheduling, routing, and collision-free navigation of AGVs inside a warehouse constitute a sequential decision-making problem that we model as a cooperative multi-agent Markov Decision Process (MDP). 
The warehouse AGV system is formulated as the tuple:

\[
\mathcal{M} = (S, A, P, R),
\]
where the components are defined below.
\paragraph{a. State Space} 
Operating in a structured warehouse with static infrastructure and dynamic obstacles (such as other AGVs), each agent uses central control signals to perceive its global environment. At any given time $t$, the AGV's state is defined by four components: warehouse layout and obstacle data ($s^{\mathrm{env}}_t$), its own operational status ($s^{\mathrm{op}}_t$), order details ($s^{\mathrm{ord}}_t$), and external factors like traffic and customer priority ($s^{\mathrm{ext}}_t$). The global state at time $t$ is expressed as:
\begin{equation}
s^{in}_t = 
\big(
s^{\mathrm{env}}_t,\;
s^{\mathrm{op}}_t,\;
s^{\mathrm{ord}}_t,\;
s^{\mathrm{ext}}_t\big)
\end{equation}

The environmental state $s^{\mathrm{env}}_t$ reaches the geometric and dynamic configuration of the warehouse as follows:
\begin{equation}
s^{\mathrm{env}}_t =
\big(
\mathcal{G},\;
\mathcal{O}_t,\;
\mathcal{B}
\big),
\end{equation}
where $\mathcal{G}$ is the warehouse layout graph (static shelves, racks, aisles, and conveyor belt), $\mathcal{B}$ are static barriers, and $\mathcal{O}_t$ are dynamic obstacles, including the real-time positions and orientations of all AGVs. For the AGV operating state $s^{\mathrm{op}}_t$, each $i$ has an internal operating state:
\begin{equation}
s^{\mathrm{op}}_{t,k} = 
\big(
\ell_t^k,\;
\theta_t^k,\;
\mathrm{load}_t^k,\;
\mathrm{battery}_t^k,\;
\mathrm{status}_t^k
\big),
\end{equation}
contains its position $\ell_t^k$ on the warehouse grid, orientation $\theta_t^k$, load status $\mathrm{load}_t^k$ (number of items), remaining battery level $\mathrm{b}_t^k$, and operational status $\mathrm{status}_t^i$ (e.g. idle, moving, picking, delivering, charging). Orders $s^{\mathrm{ord}}_t$ include pickups from shelves and deliveries to workstations or conveyor belts:
\begin{equation}
s^{\mathrm{ord}}_t =
\big(
\ell_j^{\mathrm{pick}},\;
\ell_j^{\mathrm{del}},\;
o_j,\;
\mathrm{c}_j
\big),
\end{equation}
where $\ell_j^{\mathrm{pick}}$ and $\ell_j^{\mathrm{del}}$ denote the pickup and delivery locations, respectively, $o_j$ is the release time at which the order becomes available. To integrate the last-mile delivery influence and LAN interactions, we include $s^{\mathrm{ext}}_t$:
\begin{equation}
s^{\mathrm{ext}}_t =
\big(
f_u^{\mathrm{Pareto}},\;
T_u^{\mathrm{dead}}
\big),
\end{equation}
where $f_u^{\mathrm{Pareto}}$ denotes the order(customer) Pareto-front priority for downstream routing. We have already performed Pareto's multi-objective optimization when we input this state information; therefore, no other additional information is needed.

\paragraph{b. Action Space}

From the perspective of a two-dimensional plane, the actions of AGVs $ \{\mathcal{A}_i\}$ include moving forward, backward, left, right, and staying. The AGVs can move vertically and horizontally simultaneously. After the action occurs, the AGV status is updated, a new status is generated, and the process continues. 

\paragraph{c. Transition Function}

The transition function for AGVs is deterministic as follows:
\[
P(s_{t+1} \mid s_t, a_t) = 1.
\]
This transition function states that AGV motion inside the warehouse is fully deterministic; once a state $s_t$ and an action $a_t$ are given, the next state $s_{t+1}$ is uniquely determined. For each AGV $k$, its position and orientation at the next time step are updated according to the movement implied by its action, represented by the operators $\Delta(a_t^k)$ for displacement and $\Theta(a_t^k)$ for heading change. The battery level decreases based on the time of movement. When an AGV completes a pickup or delivery, the corresponding order information in the order state is updated and the order completion time is recorded. The new positions of all AGVs form the updated dynamic obstacle set, which is used to maintain a safe and collision-free navigation. Finally, external indicators, such as congestion and routing efficiency, evolve deterministically based on the new AGV configuration, ensuring that $s_{t+1}$ consistently reflects both internal and external conditions.
For each AGV $k$:
\begin{equation}
\ell_{t+1}^k = \ell_t^k + \Delta(\,a_t^k\,), \qquad
\theta_{t+1}^k = \Theta(\,a_t^k\,).
\end{equation}
Order pick or drop events update $s^{\mathrm{ord}}_t$, and the order finish time $T_j^{\mathrm{fin}}$ is recorded when unloading is complete:
\begin{equation}
T_j^{\mathrm{fin}} = t \quad \text{if AGV $i$ completes order $u$}.
\end{equation}

\paragraph{d. Reward Machine Design}
The reward mechanism adopts multiple reward machines (RMs) \citep{ToroIcarteRodrigo2022RMER} to represent distinct operational objectives during the AGV transportation. Each RM receives an abstract description of the environment and outputs a reward based on its current state. This modular structure ensures that optimization outcomes, constraints, and external conditions are encoded as independent reward components defined as:
\begin{equation}
\mathcal{R}_{PSA} = \langle U, u_0, F, \delta_u, \delta_r \rangle,
\end{equation}

where $U$ is the set of RM states with initial state $u_0$; $F$ denotes terminal states; $\delta_u : U \times P \to U$ is the state-transition function; and $\delta_r : U \times P \to \mathbb{R}$ defines the reward. Here, $P$ is a set of propositional symbols representing the key environmental conditions at time $t$. If a condition $\sigma$ satisfies a formula $\varphi$, then the RM is updated as follows:
\begin{equation}
u_{t+1} = \delta_u(u_t, \sigma_t).
\end{equation} 
Integrating RMs into the transportation model yields the Markov Decision Process with Reward Machines (MDPRM):
\begin{equation}
\mathcal{T}
= \langle S, A, p, \gamma, \mathcal{P}, L,
U, u_0, F, \delta_u, \delta_r \rangle,
\end{equation}
where $\gamma$ is the discount factor, and $\mathcal{P}$ is the set of propositions. 
At each time step $t$, the optimization outputs determine $\sigma_t$, which updates the RM state and produces the corresponding reward. 
In this study, we designed four reward machines $RM_{En}, RM_{In}, RM_{Ex}, RM_{Sa}$, each with its own state transitions ($\delta_{u}^{En}, \delta_{u}^{In}, \delta_{u}^{Ex}, \delta_{u}^{Sa}$) and reward functions ($\delta_{r}^{En}, \delta_{r}^{In}, \delta_{r}^{Ex}, \delta_{r}^{Sa}$). These track distinct dimensions of the problem: energy consumption, inventory management cost, external influence, and safety/stability. Thus, at each period $t$, each agent $i$ receives a vector of RM rewards:

\begin{equation}
R_{i,t}
= \left( R^{En}_{i,t},\ R^{In}_{i,t},\ R^{Ex}_{i,t},\ R^{Sa}_{i,t} \right).
\end{equation}


Sparse rewards encode strategic long-term objectives that are not associated with every movement of the AGV. They are triggered upon task completion events, such as package pickup or route completion. The sparse reward combines the inventory cost and external influences as follows:
\begin{equation}
R_{\text{sparse}}(t) = w_{Ex}^{\text{sparse}} \times r_{Ex}^{\text{sparse}} 
+ w_{In}^{\text{sparse}} \times r_{In}^{\text{sparse}}
\end{equation}
In contrast, instantaneous rewards evaluate the immediate impact of each action performed by the AGV. 
They represent operational factors that fluctuate with each movement:

\begin{equation}
R_{\mathrm{instant}}(t) =
w_{Sa}^{\mathrm{instant}}\, r_{Sa}^{\mathrm{instant}}
+
w_{En}^{\mathrm{instant}}\, r_{En}^{\mathrm{instant}}.
\end{equation}
Among these, reward design based on external comprehensive factors is relatively complex and requires integration to find a set of optimal solutions for multiple objectives (higher customer class, lower traffic congestion, and shorter distance). We adopted an exponential weighting scheme defined over the Pareto-layer indices. Each order is assigned to a Pareto front that characterizes its relative priority, where lower-index fronts correspond to higher service importance, and orders within the same front are treated equally. Instead of allocating rewards across all frontends according to a uniform standard (which could disproportionately penalize frontends with a large number of orders), we assign a reward to each order, which decays exponentially with its frontend index. This formulation ensures strict dominance, such that all orders in higher priority fronts receive strictly greater rewards than those in lower priority fronts. By specifying the boundary conditions for the highest and lowest fronts, the exponential structure enables explicit control over the spacing between priority levels while maintaining a smooth and tunable reward mapping. Let 
$f \in \{f_{\min}, \dots, f_{\max}\}$ denotes the Pareto front index, with a lower $f$ indicating a higher priority. Let $F_{\max}$ denote the maximum front index in the current decision window and $r(f)$ the sparse reward for orders in front $f$. The priority reward function is defined as:

\begin{equation}
r_{ex}(f) = K \cdot b^{\,F_{\max} - f}, \qquad b > 1, \; K > 0,
\end{equation}
where $b$ and $K$ are determined from the boundary conditions. Specifically, we set
\[
\begin{cases}
r_{ex}(f_{\text{high}}) = R_{\text{high}}, \\[6pt]
r_{ex}(f_{\text{low}}) = R_{\text{low}},
\end{cases}
\]
where $f_{\text{high}}$ is the best Pareto layer (e.g., $f=2$) with target reward $R_{\text{high}}$ (e.g., $20$), and $f_{\text{low}}$ is the worst Pareto layer (e.g., $f=22$) with target reward $R_{\text{low}}$ (e.g., $0.1$).
By dividing the two equations, we obtain the exponential base as follows:
\begin{equation}
\label{b}
b = \left(\frac{R_{\text{high}}}{R_{\text{low}}}\right)^{\tfrac{1}{f_{\text{low}} - f_{\text{high}}}},
\end{equation}
and the scaling constant:
\begin{equation}
K = \frac{R_{\text{high}}}{b^{\,F_{\max} - f_{\text{high}}}}.
\end{equation}

\medskip
\noindent
Thus, the final reward assignment for each order is
\begin{equation}
r_{ex}(f) = \frac{R_{\text{high}}}{b^{\,F_{\max} - f_{\text{high}}}} \cdot b^{\,F_{\max} - f},
\end{equation}
with $b$ given by (\ref{b}).

\paragraph{e. Weight Space}
At each step, all reward machines observed the environment and updated their internal states separately. The system then performs parallel tracking to compute the final reward as a composition of sub-rewards 
$R_{i,t}^{En}$, $R_{i,t}^{In}$, $R_{i,t}^{Ex}$, $R_{i,t}^{Sa}$. Because the relative importance of these sub-rewards changes over time and across scenarios, the weights determine the extent to which each factor contributes to the final decision: when a factor becomes more influential, its corresponding weight increases, and when its influence diminishes, the weight decreases or approaches zero. We define the weight space as: 

\begin{equation}
\Omega = \{ w \;|\; w = [w_1, w_2, w_3,\dots, w_k]^T, \; w_j \in [0,1] 
\end{equation}
where each $w_j$ corresponds to a sub-reward in the multi-objective reward function vector $R(s,a) = [r_1, r_2, r_3,$ $\dots, r_k]^T$. A multi-objective MDP is thus an MDP in which the reward function 
$R: S \times A \to \mathbb{R}^n$ outputs an $n$-dimensional reward vector \citep{RoijersD.M.2013ASoM}. 

\subsubsection{RL Architecture for Last-Mile Delivery on External Multimodal System}
The proposed MDP captures state transitions driven by vehicle movements, customer service events, and real-time travel time variations influenced by congestion and regional traffic factors. The heterogeneous capacitated vehicle routing is formulated as the tuple:
\[
\mathcal{M} = (S, A, P, R),
\] 
where the components are defined as follows.

\paragraph{a. State Space}
\label{sec:state_space}
Integrating external logistics data with internally optimized warehouse outputs, the heterogeneous fleet's state depends on four components: the vehicle's operational status and attributes ($s^{\mathrm{veh}}_t$), customer details and priorities ($s^{\mathrm{cus}}_t$), the dynamic transportation environment ($s^{\mathrm{traf}}_t$), and the periodically updated warehouse order sequence ($s^{\mathrm{seq}}_t$).
The state at decision step $t$ is decomposed into four components as follows:
\begin{equation}
s^{ex}_t = \big( s^{\mathrm{veh}}_t,\; s^{\mathrm{cus}}_t,\; s^{\mathrm{traf}}_t,\; s^{\mathrm{seq}}_t \big).
\end{equation}
Each vehicle state records its current location, remaining capacity, accumulated travel time and distance, partial route, speed, capacity constraints, and parameters. This fully captures the heterogeneity across vehicles. The vehicle state aggregates the status and static attributes of every vehicle $h\in\mathcal H$:
\begin{equation}
s^{\mathrm{veh}}_t = \big( \ell_t^h,\; o_t^h,\; T_t^h,\; D_t^{h},\; G_t^h,\; f_h^{\mathrm{base}},\; Q_h,\; D_h^{\max},\; z_h \big).
\end{equation}
where $\ell_t^h$ shows the current location (node) of vehicle $h$, $o_t^h$ indicates the remaining capacity, $T_t^h$ is the accumulated travel time since departure, $D_t^{h,\mathrm{cum}}$ is the accumulated distance travelled in the current route. For multiple trips, $G_t^h$ represents a partial route (ordered list of assigned nodes). For the customer(order) state:
\begin{equation}
s^{\mathrm{cust}}_t = \big(l_u,\;d_u^t,\;p_u,\;r_u,\,\;T_u^{\mathrm{dead}},\;T_u^{\mathrm{ser}},\mathrm{served}_u
\big).
\end{equation}
It includes the customer location in $l_u$, remaining demand in $d_u^t$, priority class in $p_u$, release time (packaging completion time) in $r_u$, deadline, service time, and service status. New orders are appended every five minutes through the warehouse sequence mechanism.
While Traffic(network) state $s^{\mathrm{traf}}_t$ can be defined in:
\begin{equation}
s^{\mathrm{traf}}_t =
\big(
\mathcal{R},\;
\beta(t),\;
\gamma(r),\;
\lambda_{hrt}
\big)
\end{equation}
where $\mathcal{R}$ illustrates the partition of the service area into regions(centre,urban, or suburb). Finally, we apply the optimized sequence from the warehouse into vehicle routing $\mathcal{Q}_t$ with an update window $\Delta_t$:

\begin{equation}
s^{\mathrm{seq}}_t =
\big(
\mathcal{Q}_t,\;
\Delta_t
\big)
\end{equation}
representing the warehouse order preparation process, queue evolution, ABC categories, and the next update window.

\paragraph{b. Action Space}

At each decision step, the agent selects vehicle $veh_h$ to serve customer $y_u$ in $\mathcal{A}(s_t)$. 

\paragraph{c. Transition Function}
The transition function $\mathcal{P}: S \times A \to S$ defines how the external fleet system evolves after each action $a_t = (veh_h, y_u) \in \mathcal{A}(s_t)$. First t,he vehicles are updated by:
\begin{align}
\ell_{t+1}^h &= l_u, & o_{t+1}^h &= o_t^h - d_u^t, \nonumber\\
T_{t+1}^h &= T_t^h + t_{travel}(\ell_t^h,l_u) + T_u^{\mathrm{ser}}, & 
D_{t+1}^{h} &= D_t^h + d(\ell_t^h, l_u), \nonumber\\
G_{t+1}^h &= G_t^h \cup \{l_u\}. & \nonumber
\end{align}
where $t_{travel}(\ell_t^h, l_u)$is the estimated travel time from the vehicle's current location $\ell_t^h$ to the next customer location $l_u$, depending on the distance, traffic, and speed. In addition, customer demand changes as follows:
\begin{equation}
d_u^{t+1} = 0, \quad \mathrm{served}_u = 1, \quad 
\end{equation}
For traffic conditions and warehouse sequences, they are updated in a specific time window so that the system records new conditions over time as a transition. This transition defines the evolution of the heterogeneous fleet system after each assignment action. The vehicle states are updated according to the travel, service, capacity, and route progression. Customer states reflect fulfilled demand and service completion, whereas unserved customers retain their previous status. Traffic and environmental conditions are updated dynamically based on congestion, regional parameters and historical statistics. The warehouse sequence is periodically refreshed to incorporate new orders, ABC prioritization, and preparation times. Feasibility constraints ensure that only actions respecting capacity, distance, time windows, and deadlines are allowed, thus guaranteeing valid transitions at every decision step. 
\paragraph{d. Reward Design}
Unlike the controlled warehouse environment, external multi-node vehicle routing on a network, without the need for continuous real-time monitoring, faces numerous uncontrollable factors. To achieve multi-objective collaborative optimization in this environment, we use a linear calculation method to assign weights to different goals. Specifically, the objective function is decomposed into three computable sub-objectives: priority satisfaction, transportation time efficiency, and congestion adaptability. At each decision step $t$, the reward vector is as follows:

\begin{equation}
\mathbf{R}_t^{\mathrm{VRP}} = \left( r_t^{\mathrm{Pr}},\; r_t^{\mathrm{Tr}},\; r_t^{\mathrm{Con}} \right)
\label{eq:reward_vector}
\end{equation}
where $r_t^{\mathrm{Pr}}$ reflects the responsiveness to high-priority customers, $r_t^{\mathrm{Tr}}$ measures transit time, and $r_t^{\mathrm{Con}}$ assesses the impact of current traffic conditions. The priority reward $r_t^{\mathrm{Pr}}$ considers the relative priority differences among customers as follows:

\begin{equation}
r_t^{\mathrm{Pr}} = -\rho_p \cdot \left(w_{\max} - w_u\right)
\label{eq:priority_reward}
\end{equation}

where $w_u$ is the priority weight of the current customer $u$, $w_{\max} = \max_{v \in \mathcal{U}_t} w_v$ represents the highest priority of unserved customers, and $\rho_p > 0$ is the priority penalty coefficient. The transportation time reward $r_t^{\mathrm{Tr}}$ is calculated based on the vehicle base speed and estimated distance as follows:

\begin{equation}
r_t^{\mathrm{Tr}} = -\gamma \cdot \frac{D_{uv}}{v_h^{\mathrm{base}}}
\label{eq:transport_reward}
\end{equation}

where $D_{uv}$ denotes the distance from the vehicle’s current position to target customer $u$, $v_h^{\mathrm{base}}$ represents the average or base speed of vehicle $h$ for heterogeneous vehicles, and $\gamma > 0$ is the time-related penalty coefficient. The congestion reward $r_t^{\mathrm{Con}}$ comprehensively considers the effects of time, region, and vehicle type on travel efficiency:

\begin{equation}
r_t^{\mathrm{Con}} = -\kappa \cdot \big[ (1 - \beta_h(t)) + (1 - \gamma_h(r)) + \lambda_{hrt} \big]
\label{eq:congestion_reward}
\end{equation}

where $\beta_h(t) \in (0,1]$ reflects the impact of time period(normal time or busy time) ; $\gamma_h(r) \in (0,1]$ represents regional influence factor(metropolitan, urban, or suburb) which characterizes the inherent congestion level of region $r$; $\lambda_{hrt} \in [0,1)$ shows congestion propagation factor,influencing the congestion diffusion in region $r$ at time $t$; $\kappa > 0$ the congestion penalty coefficient.

The total reward within each time window is computed as the product of the individual reward components:

\begin{equation}
R_t^{\mathrm{total}} = r_t^{\mathrm{Pr}} \cdot r_t^{\mathrm{Tr}} \cdot r_t^{\mathrm{Con}}
\label{eq:total_reward}
\end{equation}

The multiplicative formula for the reward ensures that all reward components must be satisfied simultaneously to obtain a higher total reward, which aligns with our initial intention of multi-objective optimization.

\subsection{Interactions in Integrated Internal/External Systems}
\begin{figure}[!ht]
\centering
  \includegraphics[width=1.1\textwidth]{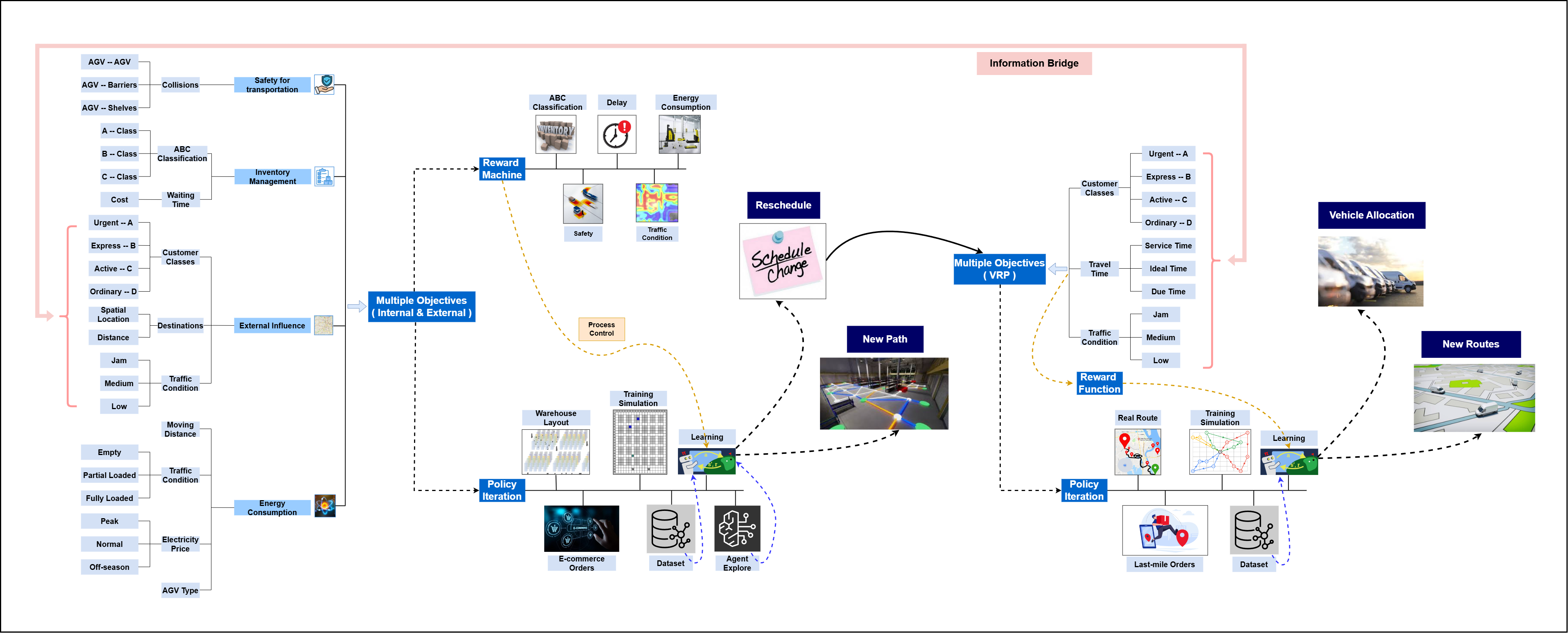}
  \caption{Integration Optimization Framework}\label{fig:integration}
\end{figure}

To achieve integrated optimization, an interactive bridge must be built between the mechanisms of the two systems inside and outside the warehouse. That is, we want to ensure that the scheduling and path planning optimization within the warehouse system considers the influence of external factors, and that the allocation and routing optimization of the last kilometre of transportation outside the warehouse is based on the order within the warehouse, as shown in Figure \ref{fig:integration}.


As multi-objective optimization requires balancing trade-offs rather than optimizing all goals simultaneously \citep{alma992323105405151,KimSejin2022Mmou}, calculating high-dimensional Pareto fronts is often computationally expensive and difficult to interpret. To overcome this, we used a 3D Pareto front to translate external factors into time-based AGV picking sequences, which then served as inputs for our neural network.
High-level Pareto optimization focuses on the concepts of energy consumption, inventory control in warehouses, external influence, and safety assurance, and applies these results to the design of specific reward formulas in reward machines. For linear optimization, such as energy consumption and inventory cost for orders, existing research results have proven the feasibility of linear programming (LP), which is usually solved using mature methods and tools \citep{YangRunzhe2019AGAf}. For nonlinear or non-convex optimization, such as external influence factors and the final multi-objective optimization function, Pareto optimization or Pareto front is particularly powerful for solving problems that are not scalable or not at the same unit level in multi-objective optimization without convexity assumptions \citep{KayaC.Yalçın2023OotP}.
Similarly, in the process of providing delivery services using external fleets, we employ a multi-objective optimization approach. During vehicle and customer selection, we comprehensively weighed the total travel time and priority service capabilities. By establishing a multi-objective mechanism, we reward participants for achieving each small milestone, thereby guiding heterogeneous fleets to allocate resources and serve customers.

\subsection{Framework of Integrated Optimization}
\subsubsection{Inner Warehouse Framework}
\begin{figure}[!ht]
\centering
  \includegraphics[width=1.0\textwidth]{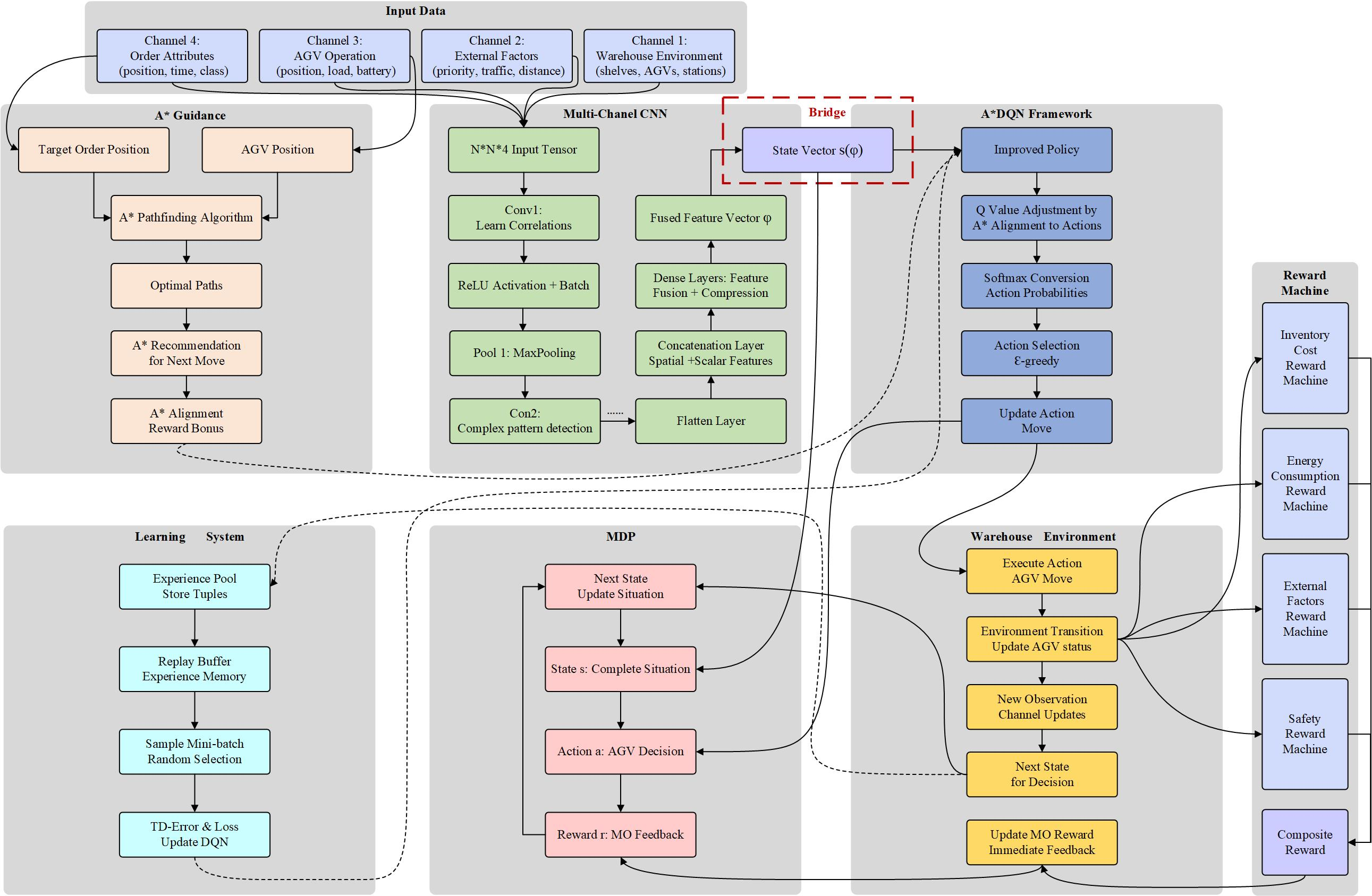}
  \caption{Inner Warehouse MORL Framework}\label{fig:CNN-Inner-framework}
\end{figure}

We propose a comprehensive MORM-AGDQN framework (Figure \ref{fig:CNN-Inner-framework}) that integrates classical planning algorithms with modern deep reinforcement learning to create an intelligent control system for Autonomous Guided Vehicles (AGVs) in dynamic warehouse environments. The system architecture comprises four fundamental components: (1) a multichannel state representation module, (2) a hierarchical feature extraction network, (3) an A*-guided decision-making framework, and (4) a multi-objective reward optimization system. This hybrid architecture enables AGVs to navigate complex warehouse environments while simultaneously optimizing multiple competing objectives.

\paragraph{a. State Representation}

The warehouse environment is encoded as a structured tensor $s_t \in \mathbb{R}^{N \times N \times C}$ at each time step $t$, where $N \times N$ represents the spatial grid, and $C=4$ denotes the feature channels. Each channel captures distinct aspects of the warehouse state.

\begin{itemize}
    \item \text{Channel 1 (Warehouse Environment)}: Static infrastructure including shelf positions, charging stations, and dynamic elements such as other AGVs' positions: $s_t^{(1)} = \{shelf_{ij}, station_{ij}, agv_{ij}\}_{N \times N}$
    
    \item \text{Channel 2 (External Factors)}: Operational constraints including order priority levels, traffic density maps, and distance transforms: $s_t^{(2)} = \{priority_{ij}, traffic_{ij}, distance_{ij}\}_{N \times N}$
    
    \item \text{Channel 3 (AGV Operational States)}: Individual AGV status including current position, battery level, and load capacity: $s_t^{(3)} = \{position_{ij}, battery_{ij}, load_{ij}\}_{N \times N}$
    
    \item \text{Channel 4 (Order Attributes)}: Task-specific information including target positions, deadlines, and priority classes: $s_t^{(4)} = \{target_{ij}, deadline_{ij}, class_{ij}\}_{N \times N}$
\end{itemize}

\begin{figure}[!ht]
\centering
  \includegraphics[width=0.8\textwidth]{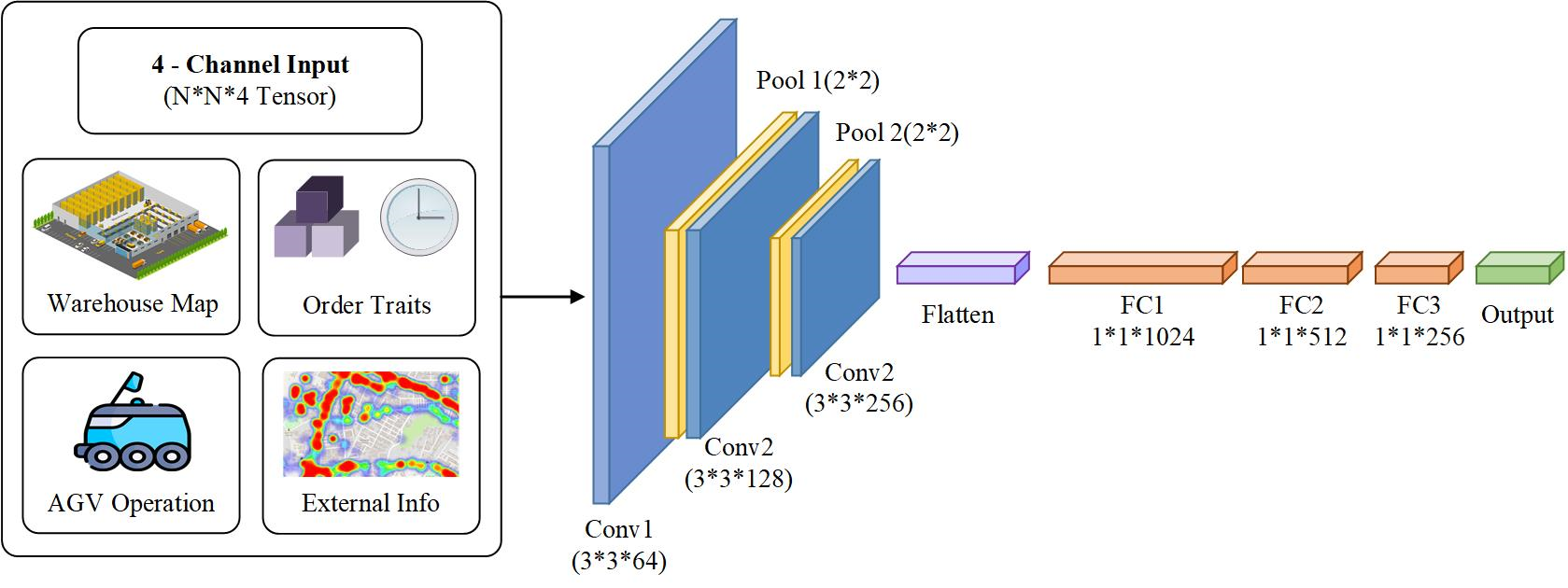}
  \caption{Inner Warehouse MORL Framework}\label{fig:CNN}
\end{figure}

This multichannel representation provides a comprehensive spatiotemporal encoding of the warehouse operational context, enabling the learning algorithm to capture the complex relationships between environmental factors and operational decisions.
\paragraph{b. Feature Extraction}
As shown in Figure \ref{fig:CNN}, the feature extraction pipeline processes a comprehensive multi-channel input state $\mathcal{S}_t \in \mathbb{R}^{N \times N \times 4}$. The input sequentially passes through three convolutional layers with 3×3 filters, Conv1 (64 filters), Conv2 (128 filters), and Conv3 (256 filters), to progressively learn low-level, mid-level, and complex global patterns. To introduce nonlinearity and stabilize training, each convolution utilizes a ReLU activation function, $\text{ReLU}(x) = \max(0,x)$, alongside Batch Normalization. Between the convolutions, 2×2 max-pooling layers (Pool1 and Pool2) reduce spatial dimensionality and provide translation invariance while preserving salient features for higher-level processing.

After flattening the hierarchical feature maps, a concatenation bridge layer integrates them with scalar features. This fused representation passes through three dense fully connected layers—FC1 (1024 units), FC2 (512 units), and FC3 (256 units), each utilizing ReLU activations and batch normalization for feature compression and training stability. This pipeline yields a distilled, fixed-length state vector, defined as $q_t = f_{\text{CNN}}(\mathcal{S}_t; \theta_c)$, where $f_{\text{CNN}}$ compresses the spatial and scalar channels into a $d$-dimensional representation ($q_t \in \mathbb{R}^d$). Finally, the MDP state transitions from $\phi$ to $s$.

\paragraph{c. A* Pathfinding and Guidance}
The distilled state vector $(q)$ is then processed by the A*-DQN Framework. Here, the framework uniquely integrates the A* pathfinding algorithm to provide an *A guided reward bonus, which steers the learning process towards globally efficient paths. The A* algorithm $\pi_{\text{A}^*}$ computes an optimal path $\tau_t^*$ from the current position of the AGV $\mathbf{p}_t$ to its target $\mathbf{g}_t: \tau_t^* = \pi_{\text{A}^*}(\mathbf{p}_t, \mathbf{g}_t)$. The recommended next action from A* is $a_t^*$, the first move in the path $\tau_t^*$. The agent selects an action using an $\epsilon$-greedy policy as follows: The Q-function $Q(q_t, a; \theta)$ estimates the expected cumulative reward. The action selection policy $\pi$ is an $\epsilon$-greedy policy enhanced with A* guidance:
\begin{equation}
\pi(a | q_t) =
\begin{cases}
\arg\max_{a'} Q(q_t, a'; \theta) & \text{P= } 1-\epsilon \\
a_t^* & \text{P=} \epsilon_{\text{A}^*} \\
\text{random } a \in \mathcal{A} & \text{P=} \epsilon - \epsilon_{\text{A}^*}
\end{cases}
\end{equation}
where $\mathcal{A}$ is the action space, and $\epsilon_{\text{A}^*}$ biases exploration toward A*-recommended actions.
\paragraph{d. Iteration of optimization}
The learning process employs experience replay with prioritized sampling to break the temporal correlations. The loss function incorporates both the value estimation error and policy regularization:
\begin{equation}
\mathcal{L}(\theta) = \mathbb{E}_{(q,a,r,q') \sim \mathcal{D}} \left[ \left( r + \gamma \max_{a'} Q(q', a'; \theta^-) - Q(q, a; \theta) \right)^2 \right] + \beta \cdot \Omega(\theta)
\label{eq:loss}
\end{equation}
where $\Omega(\theta)$ is a regularization term that penalizes large deviations from the A* policy and $\beta$ controls the regularization strength. The gradient update is performed using adaptive moment estimation as follows:
\begin{equation}
\theta \leftarrow \theta - \alpha \cdot \frac{m_t}{\sqrt{v_t + \delta}}
\label{eq:adam}
\end{equation}
where $m_t$ and $v_t$ are the first and second moment estimates, respectively, and $\delta$ is a small constant for numerical stability.
The framework also implements a dynamic exploration schedule, where $\epsilon$ and $\epsilon_{\text{A}^*}$ decay exponentially during training:
\begin{equation}
\epsilon(t) = \epsilon_{\text{min}} + (\epsilon_{\text{max}} - \epsilon_{\text{min}}) \cdot e^{-t/\tau}
\label{eq:epsilon_decay}
\end{equation}
This ensures that the agent transitions smoothly from exploration to exploitation while maintaining a bias toward A*-guided actions throughout the learning process.

\subsubsection{External Vehicle Routing Framework}
\begin{figure}[!ht]
\centering
  \includegraphics[width=1.0\textwidth]{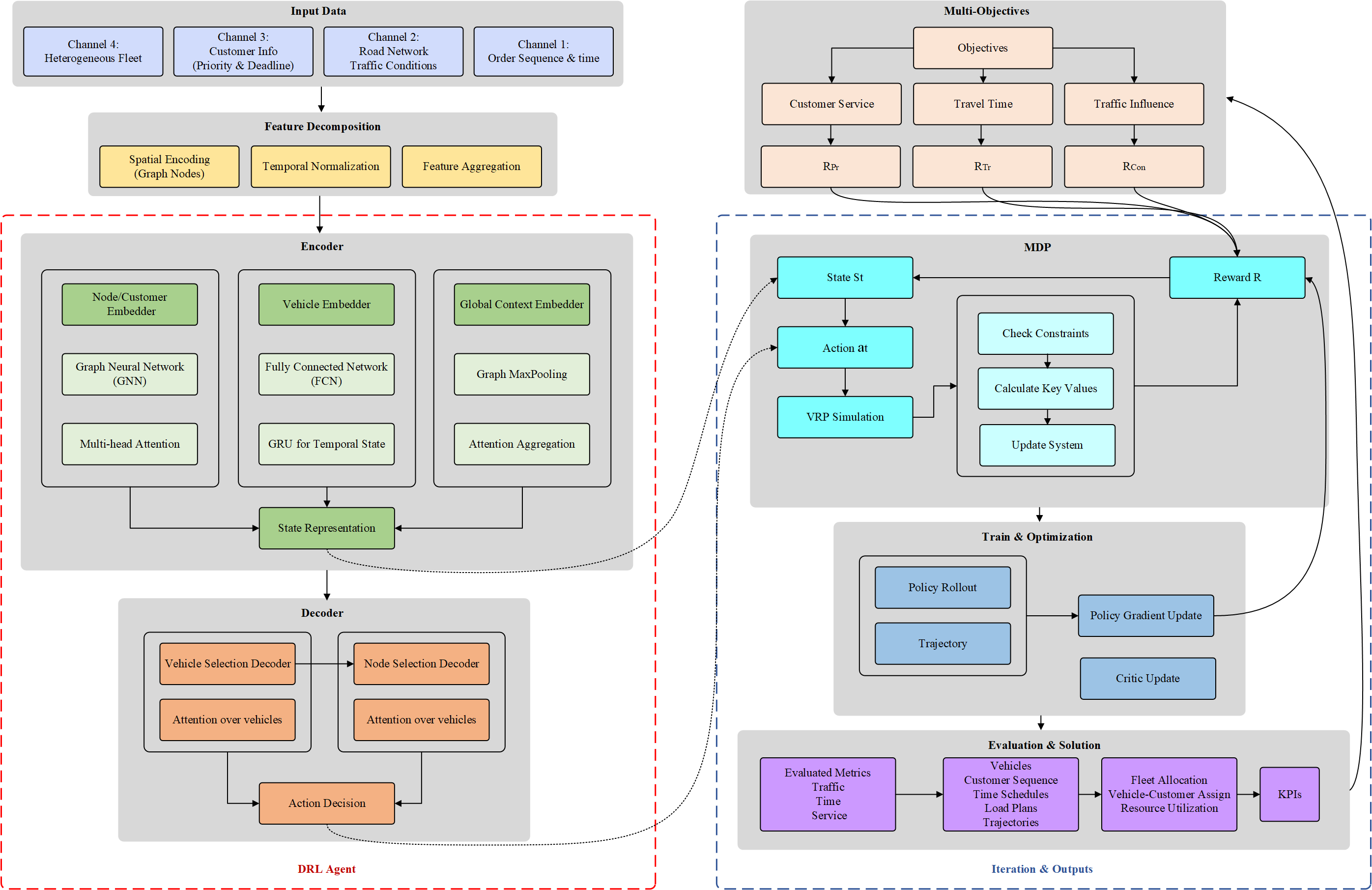}
  \caption{External Vehicle Routing Framework}\label{fig:vrp}
\end{figure}

The transportation strategy of an external heterogeneous fleet is mainly determined by the coordination of multi-objective optimization schemes based on the parcel distribution order and time within the warehouse. Its policy network $\pi_\theta(a_t|s_t)$ serves as the core guide for last-mile transportation, transforming complex states into action probability distributions. It is built on a transformer architecture consisting of three main components: an encoder for feature extraction, a vehicle-selection decoder for fleet management, and a node-selection decoder for customer allocation. The generalized framework is illustrated in Figure \ref{fig:vrp}.

\paragraph{a. Enhanced Encoder Design}
The encoder processes static problem features using specialized attention heads for transportation time, service priority, and regional traffic delay. The input representation integrates all the relevant features as follows:
\begin{equation}
\tilde{x}^u = \left(s^u, \frac{d^u}{\mathcal{Q}^1}, \frac{d^u}{\mathcal{Q}^2}, \ldots, \frac{d^u}{\mathcal{Q}^m}, w^{pr}_u, \tau_u, \frac{T^{dead}_u}{T_{max}}, \frac{T^{ideal}_u}{T_{max}}\right)
\label{eq:enhanced_input}
\end{equation}
where $s^u \in \mathbb{R}^2$ represents coordinates of customer $u$, $\frac{d^u}{\mathcal{Q}^h}$ shows demand normalized by vehicle $h$'s capacity for capacity-awareness decision, $w^{pr}_u$ is priority weight, $\tau_u \in [0,1]$ is the regional congestion coefficient for customer $u$'s zone, and $\frac{T^{dead}_u}{T_{max}}, \frac{T^{ideal}_u}{T_{max}}$ are normalized temporal constraints. The encoder employs $N$ identical layers, each containing multi-head attention (MHA) and feedforward (FF) sublayers with residual connections and batch normalization. For each head $y \in \{1,\ldots,n\}$ which represents the main objectives and key constraints for learning complex dependencies, it computes:
\begin{equation}
\begin{aligned}
Q_{l,y} &= h_l W^Q_{l,y}, \quad K_{l,y} = h_l W^K_{l,y}, \quad V_{l,y} = h_l W^V_{l,y} \\
Z_{l,y} &= \text{softmax}\left(\frac{Q_{l,y}K_{l,y}^T}{\sqrt{d_k}}\right) V_{l,y} \\
\text{MHA}(h_l) &= \text{Concat}(Z_{l,1}, \ldots, Z_{l,n}) W^O_l.
\end{aligned}
\end{equation}
The outputs from all heads are concatenated and projected as $\text{MHA}(h_l)$, enabling the model to simultaneously attend to the geographic proximity, priority relationships, temporal constraints, capacity utilization, efficiency, and traffic patterns. This comprehensive encoding provides a robust foundation for subsequent routing decision-making.
\paragraph{b. Route Feature Embedding}
The route feature embedding component captures the historical routes by encoding partially constructed routes. For each vehicle $v^h$ in a heterogeneous fleet, the algorithm extracts the corresponding node embeddings for all customers from the encoder output of its current partial path $G_{t-1}^h = [g_0^h, g_1^h, \ldots, g_{t-1}^h]$, forming the route context $\tilde{C}_t^h $:
\begin{equation}
\tilde{C}_t^h = [h_N^{g_0^h}, h_N^{g_1^h}, \ldots, h_N^{g_{t-1}^h}] 
\end{equation}
Max pooling is then applied to each $\tilde{C}_t^h$ to extract the most salient features while processing variable sequences. The pooled representations of all vehicles are concatenated into a fleet route $\tilde{C}_t^R = [\tilde{C}_t^1, \tilde{C}_t^2, \ldots, \tilde{C}_t^m]$, and then transformed through a layer with a ReLU activation function to generate the final route feature embedding as follows:
\begin{equation}
H_t^R = \text{ReLU}(\tilde{C}_t^R W_2 + b_2) 
\end{equation}
where $W_2 \in \mathbb{R}^{128 \times 512}$ and $b_2 \in \mathbb{R}^{512}$. 
This embedding preserves important historical information such as customer service details, geographic coverage, and resource utilization patterns, enabling the policy to make globally consistent routing decisions.

\paragraph{c. Vehicle Selection Decoder}

The vehicle selection decoder distributes customers to the appropriate vehicles by integrating the real-time status with multi-objective optimization requirements. The decoder constructs the vehicle feature context as follows:
\begin{equation}
C_t^{V} = [\tilde{g}_{t-1}^1, T_{t-1}^1, o_{t-1}^1, \tau_{zone}^1, \tilde{g}_{t-1}^2, T_{t-1}^2, o_{t-1}^2, \tau_{zone}^2, \ldots]
\end{equation}
which captures each vehicle's current location $\tilde{g}_{t-1}^h$, accumulated time $T_{t-1}^h$, remaining capacity $o_{t-1}^h$, and current zone congestion $\tau_{zone}^h$. This context is processed to generate vehicle embeddings as follows:

\begin{equation}
H_t^V = \text{ReLU}(C_t^V W_1 + b_1) 
\end{equation}
The decoder then combines these current status embeddings with the historical route features $H_t^R$ through concatenation and linear projection as follows: 
\begin{equation}
H_t = [H_t^V, H_t^R] W_3 + b_3 \in \mathbb{R}^{m}
\end{equation}
Finally, the vehicle distribution is:
\begin{equation}
p_t = \text{softmax}({H}_t) = \left[\frac{\exp({H}_t^h)}{\sum_{h'=1}^m \exp({H}_t^{h'})}\right]_{h=1}^m
\end{equation}
This design enables a balanced consideration of transportation time, service priority, and regional congestion during vehicle assignment.

\paragraph{d. Node Selection Decoder}
The node selection decoder, conditioned on the selected vehicle $v^h$, determines the next customer to serve by evaluating the multi-objective compatibility. The decoder constructs a context vector as follows:
\begin{equation}
H_t^c = [\bar{h}_N, h_{t-1}^h, o_t^h, \frac{T_{t-1}^h}{T_{max}}, \tau_{zone}^h, w^{avg}_t]
\end{equation}
that integrates global graph embedding $\bar{h}_N$, vehicle's last visited node embedding $h_{t-1}^h$, remaining capacity $o_t^h$, normalized accumulated time, current zone congestion $\tau_{zone}^h$, and average priority $w^{avg}_t$ of remaining customers.

\begin{figure}[h]
\centering
  \includegraphics[width=1.0\textwidth]{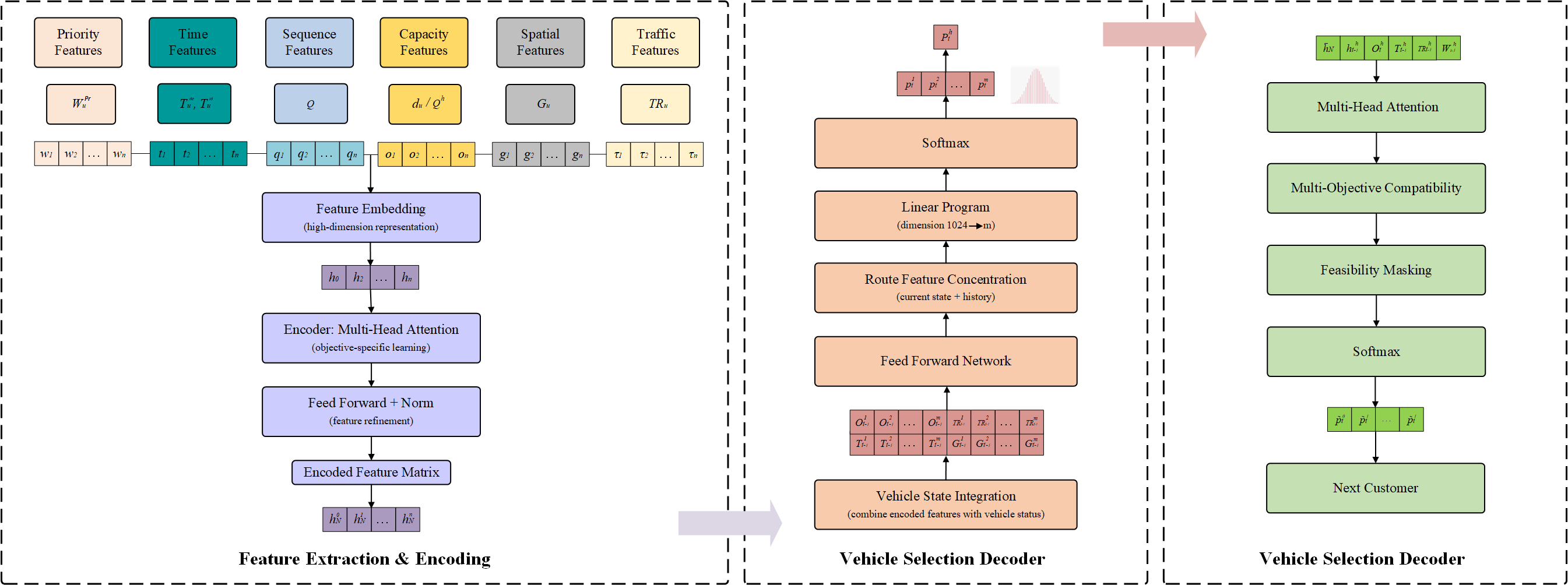}
  \caption{Multi-Head for Selection}\label{fig:vrp3}
\end{figure}

This context undergoes a multi-head attention glimpse operation as follows: 
\begin{equation}
\hat{H}_t^c = \text{MHA}(H_t^c W_c^Q, h_N W_c^K, h_N W_c^V) 
\end{equation}
This refines the representation by attending to all node embeddings. The compatibility scores between the enhanced context and each customer node are computed as follows:
\begin{equation}
u_t = \eta \cdot \tanh\left( \frac{(\hat{H}_t^c W_{comp}^Q)^T (h_N W_{comp}^K)}{\sqrt{d_k}} \right) 
\end{equation}
where the scaling factor of $\eta$ controls the entropy of the resulting distribution. Finally, a masked softmax operation generates the customer selection probabilities $\tilde{p}_t$ which is equal to $\text{MaskedSoftmax}(u_t)$, which excludes already served customers to ensure solution feasibility. 
Figure \ref{fig:vrp3} shows the process from feature extraction to vehicle and customer selection.

\section{Case Study and Simulations}\label{casestudy}
\subsection{Data Description}
To empirically test the proposed methodology, this study utilizes Amazon delivery data as the base \citep{alianwar_amazon_delivery_2024}, which aligns with standard transportation research needs. The dataset contains delivery records from last-mile logistics operations, providing comprehensive spatiotemporal information, including customer and warehouse locations, delivery-related time, order and customer characteristics, and operational timestamps. 

\subsection{Simulation Scenarios}
The simulation was developed in Python using discrete-event simulation principles, geographic data were processed using Google Maps, and traffic patterns were modeled based on historical data from the dataset. The system incorporates agent-based modeling, in which each agent operates autonomously and has real-time decision-making capabilities. 
Three operational scenarios are simulated to test the integration delivery system under varying conditions: (1) \text{Baseline Scenario} representing normal operations with regular order volumes, predictable traffic, and stable environmental factors; (2) \text{Peak Season Scenario} modeling high-demand periods with 2 to 5 times of order surges, complex traffic patterns, and various fleet sizes; and (3) \text{Uncertainty Scenario} incorporating stochastic disruptions including sudden congestion, emergency orders, and special days. These scenarios evaluate scalability, efficiency, and resilience, providing a comprehensive performance assessment across the operational spectrum.

\section{Results}\label{results}
This section presents three types of results: process results based on improved methods, comparison of different relevant algorithms, and comparison of key performance indicators (KPIs) for warehouse operations and last-mile delivery.

\subsection{Pareto Fronts Optimization}
\begin{figure}[!ht]
\centering

\begin{subfigure}{0.75\textwidth}
    \centering
    \includegraphics[
        width=\textwidth,
        height=0.32\textheight,
        keepaspectratio
    ]{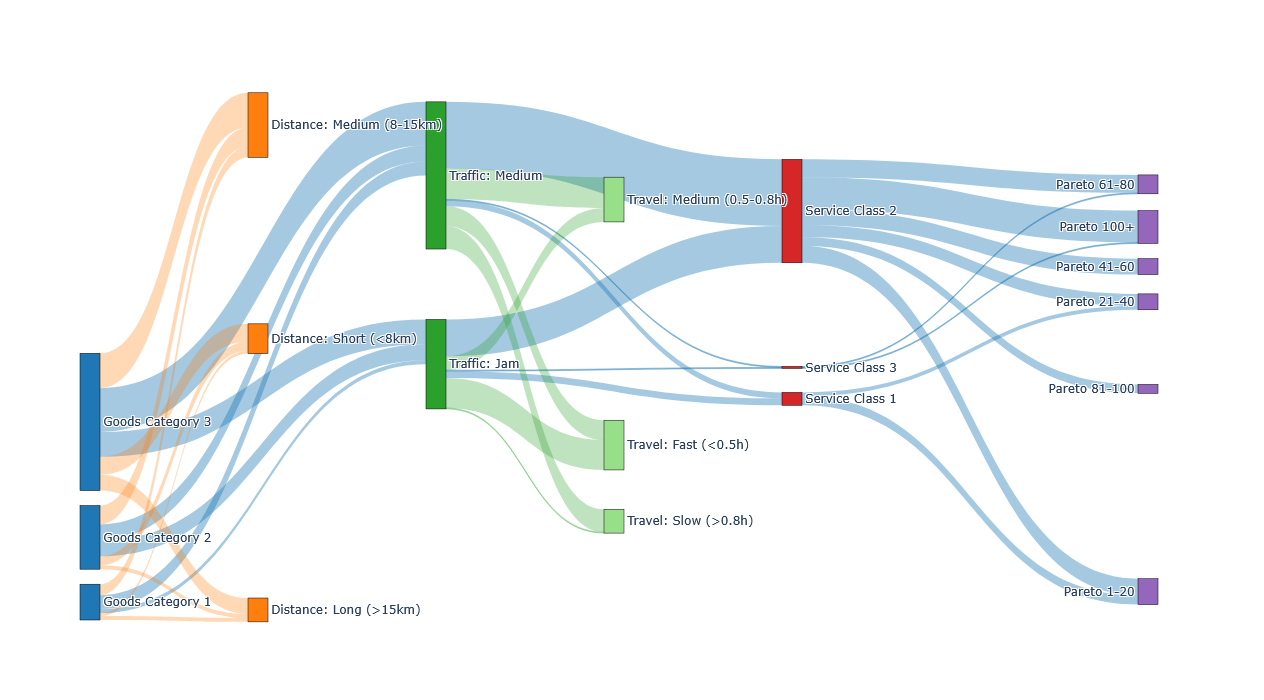}
    \caption{Normal Season}
    \label{fig:pareto_normal}
\end{subfigure}

\vspace{0.5cm}

\begin{subfigure}{0.75\textwidth}
    \centering
    \includegraphics[
        width=\textwidth,
        height=0.32\textheight,
        keepaspectratio
    ]{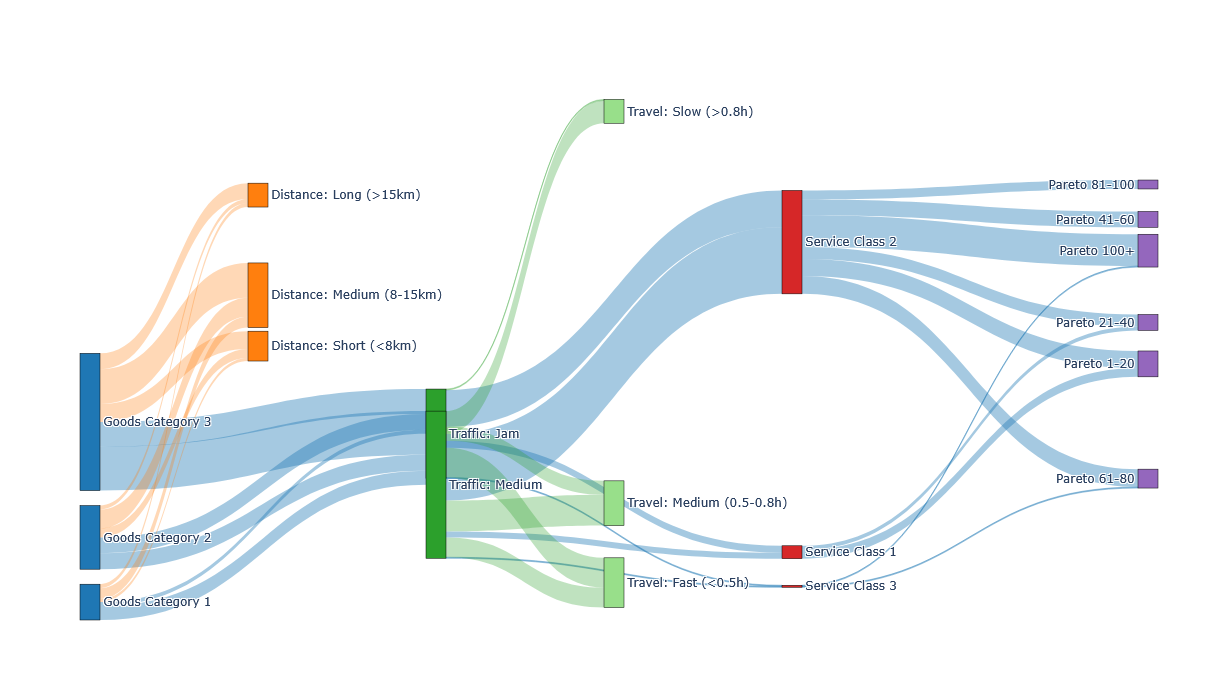}
    \caption{Peak Season}
    \label{fig:pareto_peak}
\end{subfigure}

\caption{
Multi-objective Pareto fronts optimization flow graphs. The optimization process starts from goods categories, incorporates traffic conditions and service classes, and finally converges to the Pareto front, illustrating how different operational factors influence the final multi-objective decision space.
}
\label{fig:Pareto_fronts}

\end{figure}

Figure \ref{fig:Pareto_fronts} visualizes the multidimensional relationships in logistics order dispatching, tracing the flow of orders from goods categories through traffic conditions and service classes to their final Pareto-optimal solutions. The varying widths of the streams quantitatively represent the order volumes, revealing how different operational factors converge to shape the optimized outcome space. Key patterns emerged, showing how specific combinations of product type, traffic, and service priority tended toward certain Pareto frontiers. The diagram effectively illustrates the ``funnel'' effect, where diverse starting conditions narrow through successive decision layers to reach optimal solutions, bridging complex optimization results with practical operational understanding.

\subsection{Comparison of algorithm performance}
Because the core contribution of this study is a multi-objective optimization framework rather than a computational algorithm innovation, the algorithm comparison in this section is not a traditional algorithm comparison. Instead, it aims to emphasize that our proposed multi-objective transformation and shaping strategy is more in line with actual business needs than the traditional single-objective strategy.
\subsubsection{Inner Warehouse Algorithm Comparison}
\begin{figure}[!ht]
  \centering
  \begin{subfigure}[t]{0.48\textwidth}
    \centering
    \includegraphics[width=\textwidth, height=0.35\textheight, keepaspectratio]{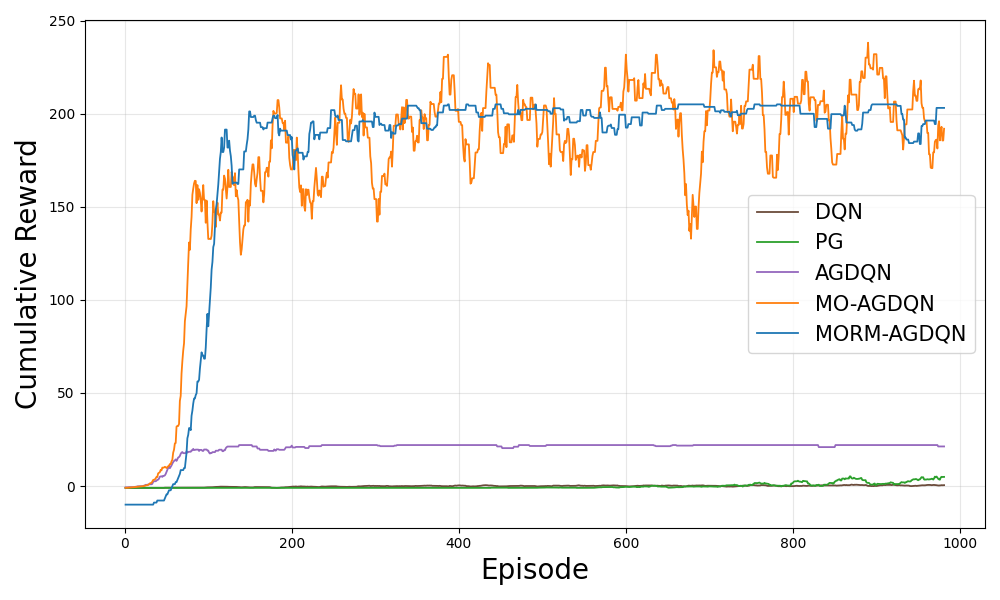}
    \caption{Inner Warehouse RL Optimization Algorithm Comparison}
    \label{fig:5-cnn algorithms}
  \end{subfigure}
  \hfill
  \begin{subfigure}[t]{0.48\textwidth}
    \centering
    \includegraphics[width=\textwidth, height=0.55\textheight, keepaspectratio]{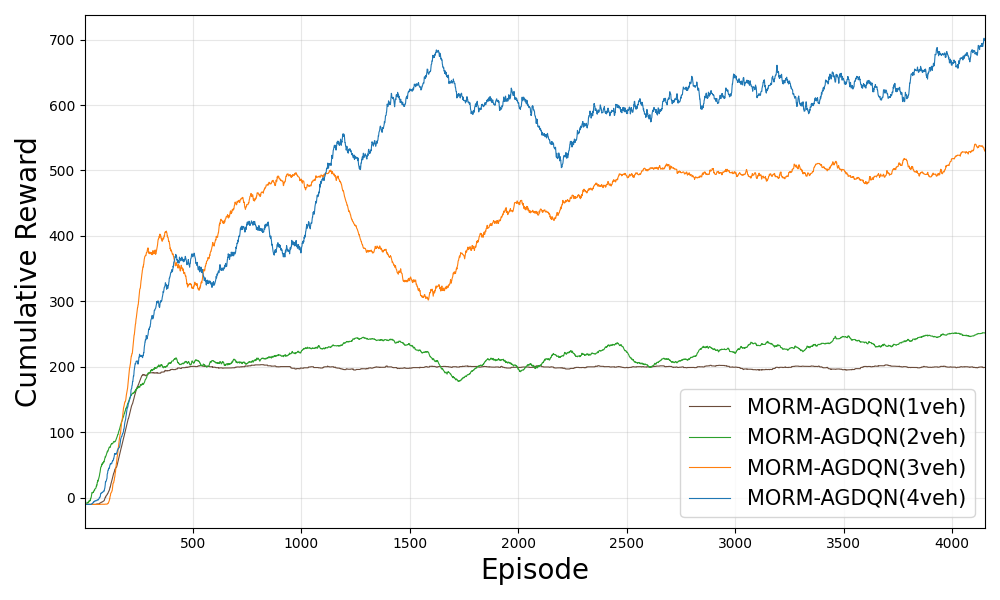}
    \caption{MORM-AGDQN Performance Analysis}
    \label{fig:multiple agv performance}
  \end{subfigure}
  \caption{Experimental MORM-AGDQN Results}
  \label{fig:CNN compare combined}
\end{figure}

For the inner warehouse algorithm, where the related paper has proved AGDQN's advantages over the PG group and Q-learning 
, we compare the rewards of the PG, DQN, AGDQN, MO-AGDQN, and MORM-AGDQN algorithms to show the advantages of our method in Figure \ref{fig:5-cnn algorithms}. For the algorithm performance, we list the MORM-AGDQN and MORM-MAGDQN algorithms for different fleet sizes and situations in Figure \ref{fig:multiple agv performance}. 

Figure \ref{fig:5-cnn algorithms} compares the cumulative reward trajectories of the five learning-based methods over 1,000 training episodes. The proposed MORM-AGDQN demonstrated distinct advantages in terms of both stability and efficacy within the inner warehouse network control context. Unlike the MO-AGDQN, which exhibits pronounced oscillations and occasional performance degradation during training, the MORM-AGDQN converges smoothly and maintains a narrow reward fluctuation band after convergence, particularly after episode 600. In contrast, the DQN, PG, and AGDQN converge to substantially lower reward levels, reflecting their limited capability to capture the complex trade-offs inherent in the transportation decision-making process. The enhanced reward profile suggests that MORM-AGDQN effectively balances exploration and exploitation, mitigates reward variance, and adapts more robustly to dynamic network conditions, leading to superior decision-making consistency and overall system resilience compared with benchmark methods.
Figure \ref{fig:multiple agv performance} demonstrates the scalability and robustness of the MORM-AGDQN algorithm as order volumes and fleet sizes increase during peak hours. While smaller scenarios (1 AGV, 17 orders) converge rapidly to a stable, lower reward, larger scenarios (up to 4 AGVs, 120 orders) ultimately achieve much higher cumulative rewards, proving the algorithm's effectiveness in complex environments. However, this performance reveals a learning trade-off: larger scales increase coordination complexity and congestion effects, resulting in longer training times and greater mid-training fluctuations before final stabilization.

\begin{figure}[!ht]
\begin{subfigure}{0.5\textwidth}
  \centering  
  \includegraphics[height=6cm]{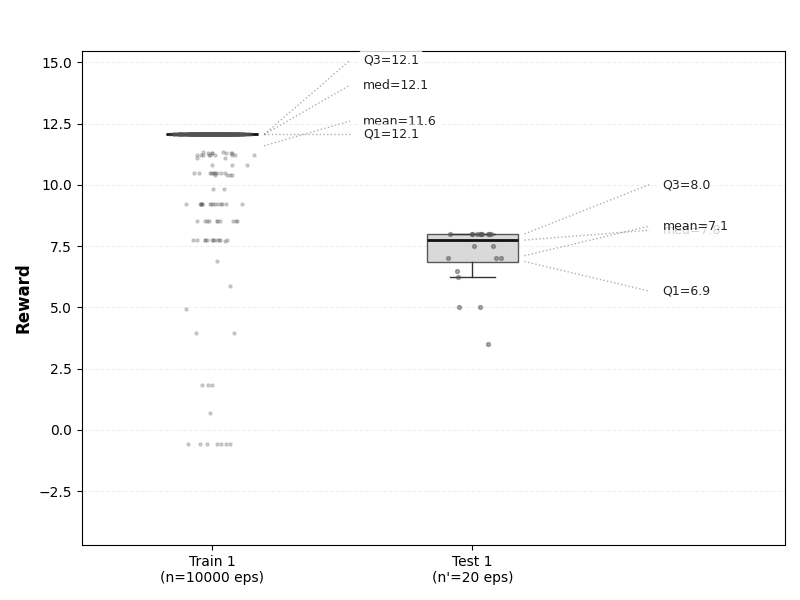}
  \caption{Small Group Size}
\end{subfigure}%
\hfill  
\begin{subfigure}{0.5\textwidth}
  \centering  
  \includegraphics[height=6cm]{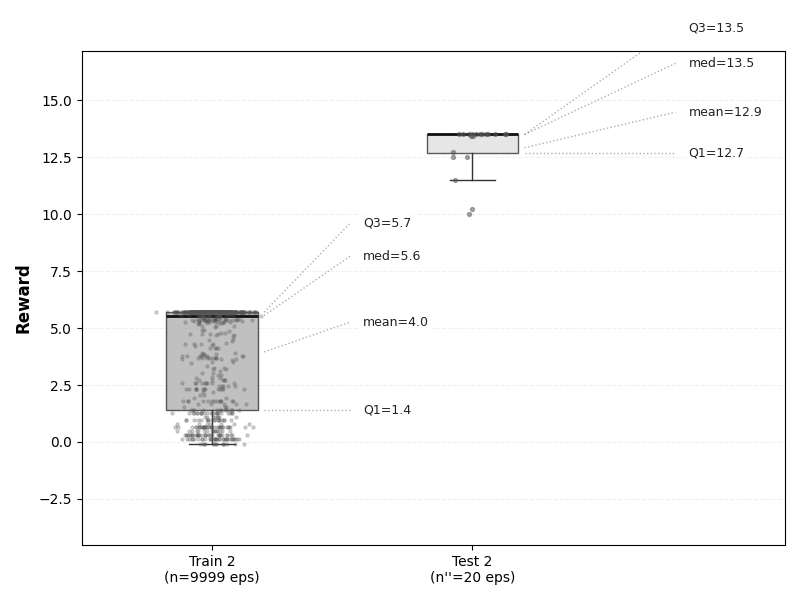}
  \caption{Large Group Size}
\end{subfigure}
\caption{Testing Average Reward of Per Order (Inner Warehouse). The two graphs have the same number of training and testing iterations, but the number of orders trained is different.}
\centering
\label{fig:box test}
\end{figure}

To verify the hybrid algorithm's generalization ability (see Figure \ref{fig:box test}), we evaluated models trained on small (17 orders) and large (96 orders) datasets. After 10,000 rounds, both models successfully generalized to complete order tasks. While small-batch training enabled AGVs to complete minor tasks with lower average rewards per order, large-batch training provided the data experience needed to adapt to environmental changes, resulting in higher average rewards. Although these average rewards eventually plateau rather than increasing indefinitely with scale, the results confirm that the framework successfully teaches the AGV to learn strategies and intelligently analyze path selections for real-world operation.

\subsubsection{External Vehicle Routing Algorithm Comparison}
Before shipping, orders in the warehouse undergo packaging and labeling confirmation and other checks. Queuing issues arise when there are a certain number of orders in the system. We followed the average processing time provided by the original data and divided this time into three parts: 5, 10, and 15 min. After training, the priority orders were processed sequentially. However, the processing capacity and throughput are limited. Therefore, the same batch of orders may be assigned to road transport vehicles, and the same batch of transport vehicles may serve multiple batches of orders. For example, Figure \ref{fig:Gant} shows the VRP problem of last-mile delivery at time node 19:10. At this time node, orders at 18:55, 19:00, and 19:05 (the system updates information every five minutes) can be delivered. These three time nodes correspond to the lowest, medium, and highest priority orders for that time period, respectively.

\begin{figure}[!h]
\begin{subfigure}{\textwidth}
  \centering
  \includegraphics[height=5cm]{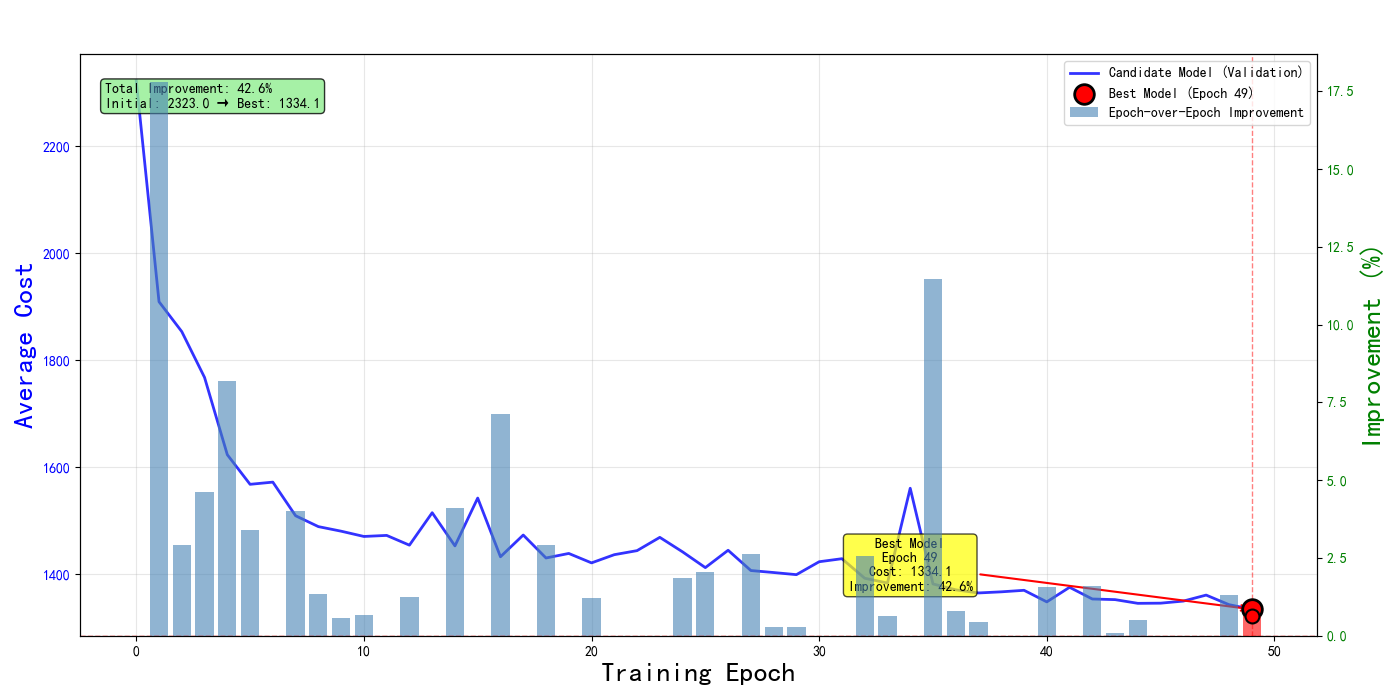}
  \caption{Scooter}
\end{subfigure}

\vspace{1em}

\begin{subfigure}{\textwidth}
  \centering
  \includegraphics[height=5cm]{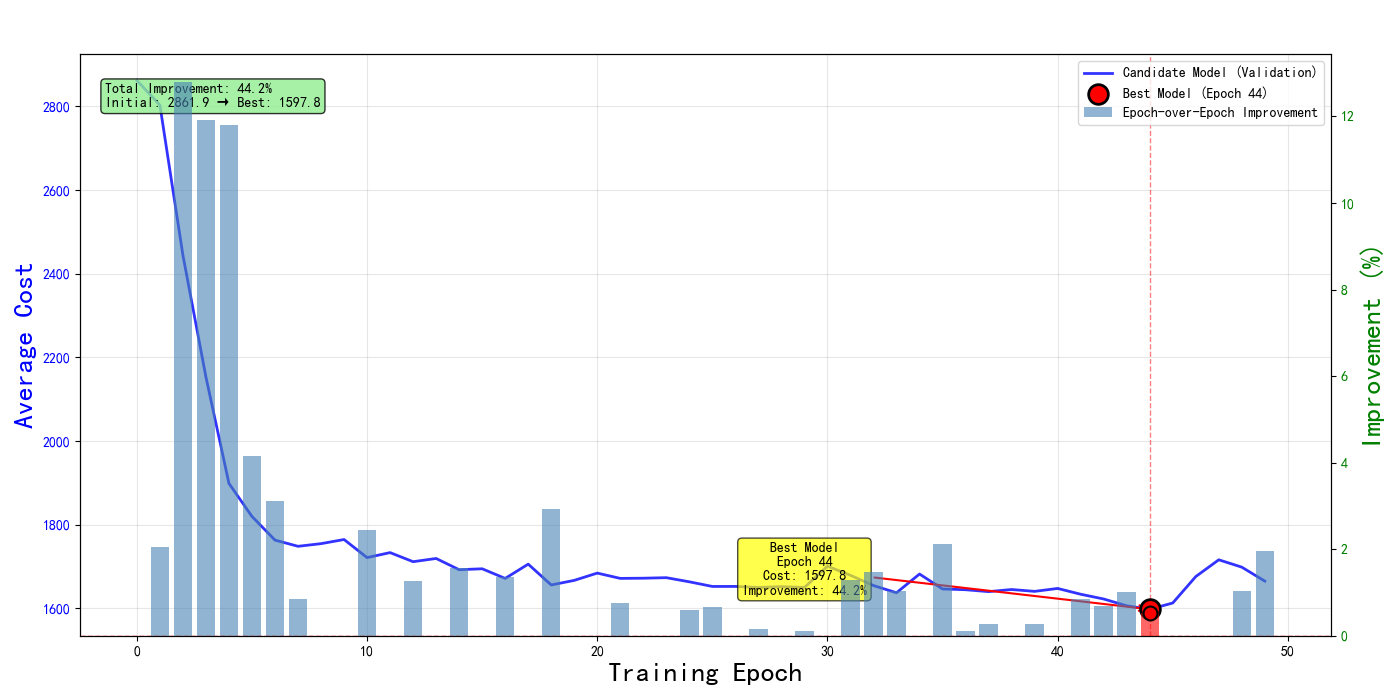}
  \caption{Motorcycle}
\end{subfigure}

\vspace{1em}

\begin{subfigure}{\textwidth}
  \centering
  \includegraphics[height=5cm]{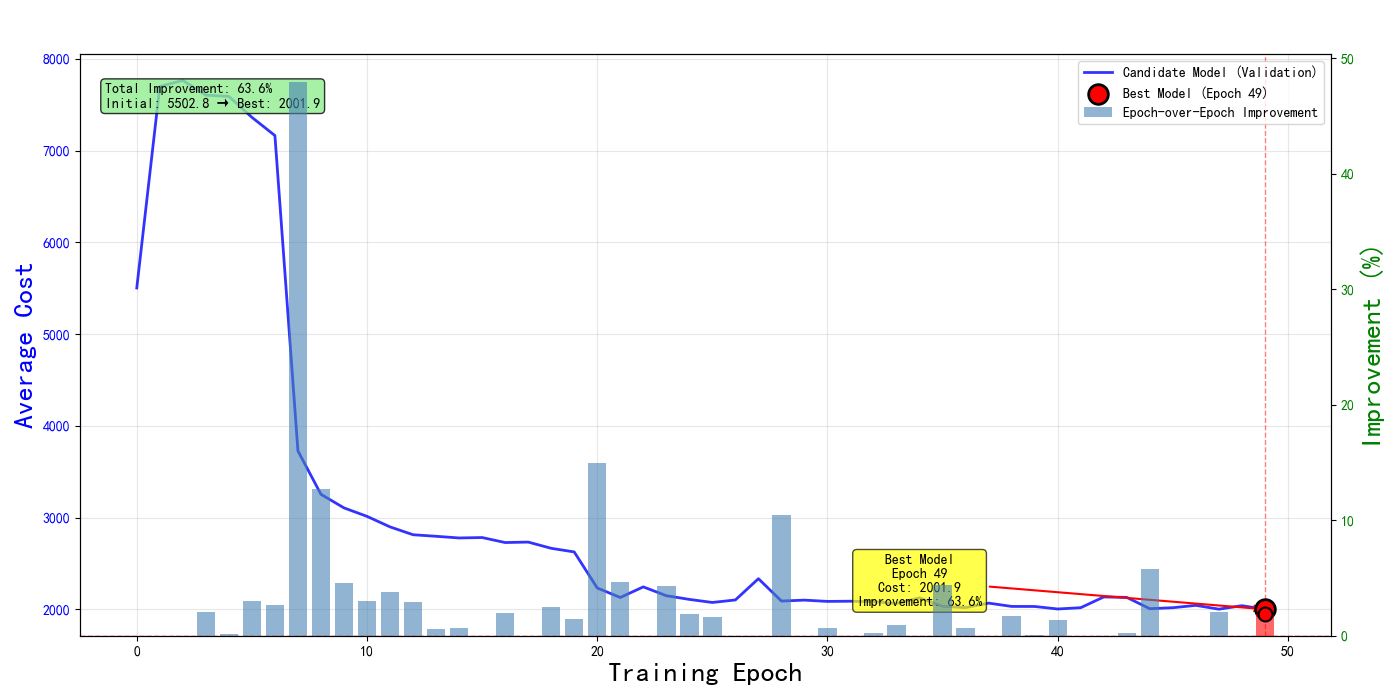}
  \caption{Van}
\end{subfigure}

\caption{Heterogeneous vehicle training convergence curve and iteration (Last-mile). The three graphs have the same number of training (50 epochs) and iterations, but the number of orders trained is different depending on travel distance.}
\label{fig:3 method routes}
\end{figure}

To prevent overfitting, we employed a rolling baseline strategy that updates the model only when it significantly outperforms historical bests on an independent test set ($p < 0.05$). In testing, our framework's distance-aware allocation, which matches vehicle capacity, speed, and battery limits to appropriate delivery ranges, achieved rapid convergence (reaching optimal solutions by iterations 44–49) and substantial cost reductions. As shown in Figure \ref{fig:3 method routes}, the method improved short-to-medium distance transport costs by 44.2\% for scooters (24 orders) and 42.6\% for motorcycles (30 orders), while boosting long-distance van transport by 63.6\% (48 orders). The model's low volatility and stable convergence confirm its robustness, adaptability, and energy efficiency across varied routing conditions.

\begin{sidewaysfigure}[p]
\centering

\begin{subfigure}[t]{0.49\textwidth}
    \centering
    \includegraphics[width=\linewidth,height=1.5\textheight,keepaspectratio]
    {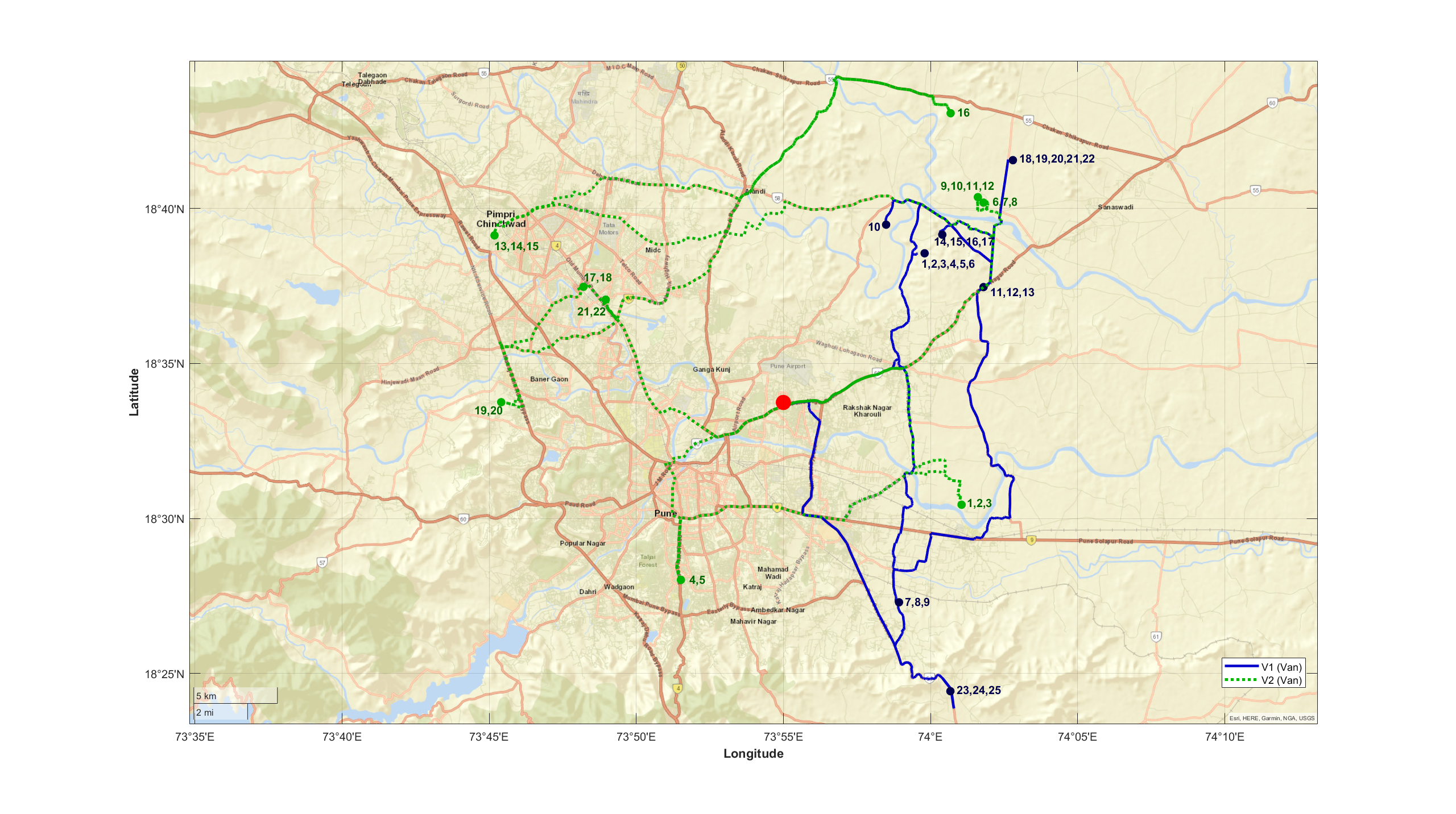}
    \caption{Time Sequence \& Clustering}
\end{subfigure}
\hfill
\begin{subfigure}[t]{0.49\textwidth}
    \centering
    \includegraphics[width=\linewidth,height=1.5\textheight,keepaspectratio]
    {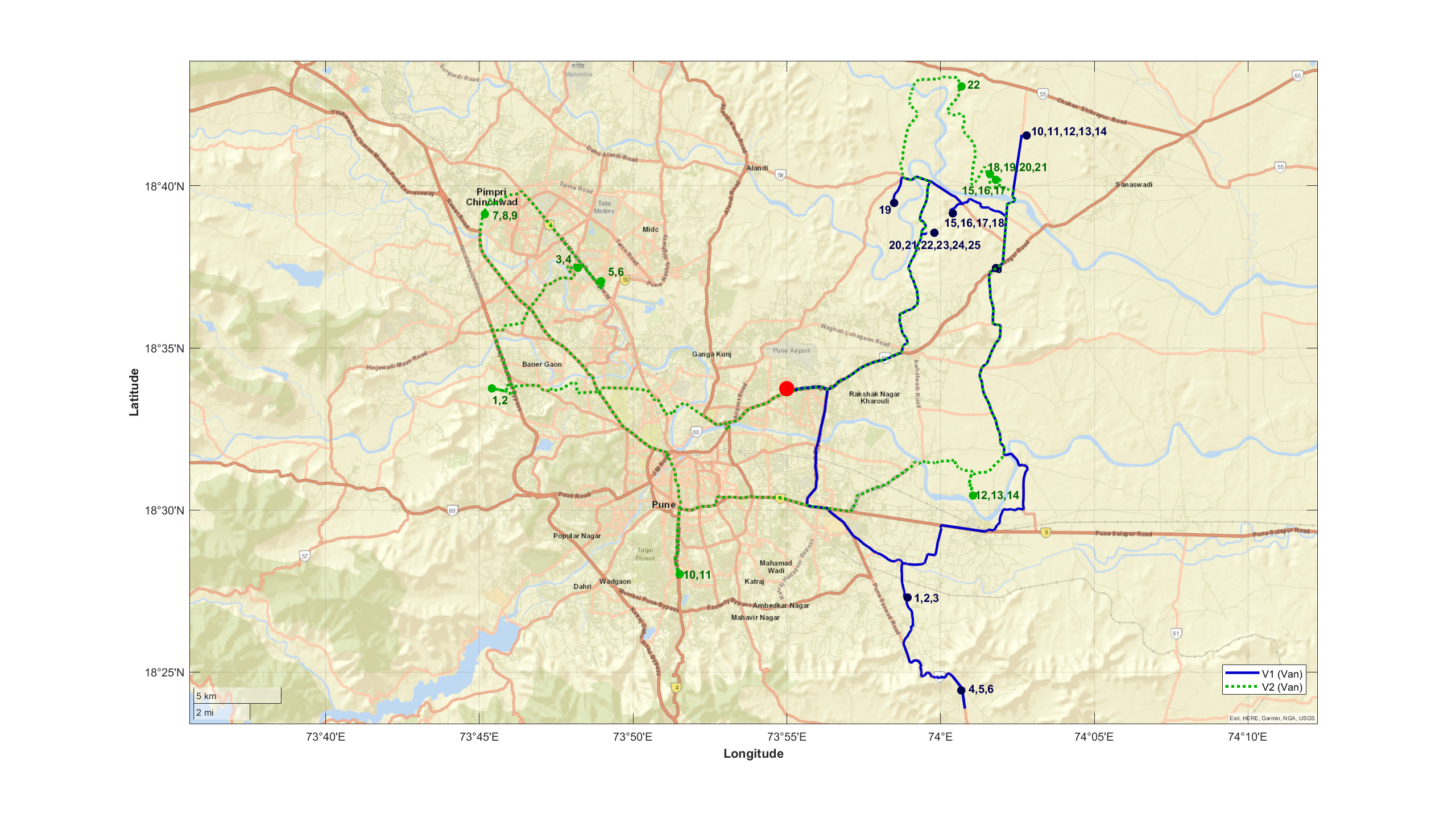}
    \caption{Shortest Path \& Clustering}
\end{subfigure}

\vspace{0.8cm}

\begin{subfigure}[t]{0.49\textwidth}
    \centering
    \includegraphics[width=\linewidth,height=1.5\textheight,keepaspectratio]
    {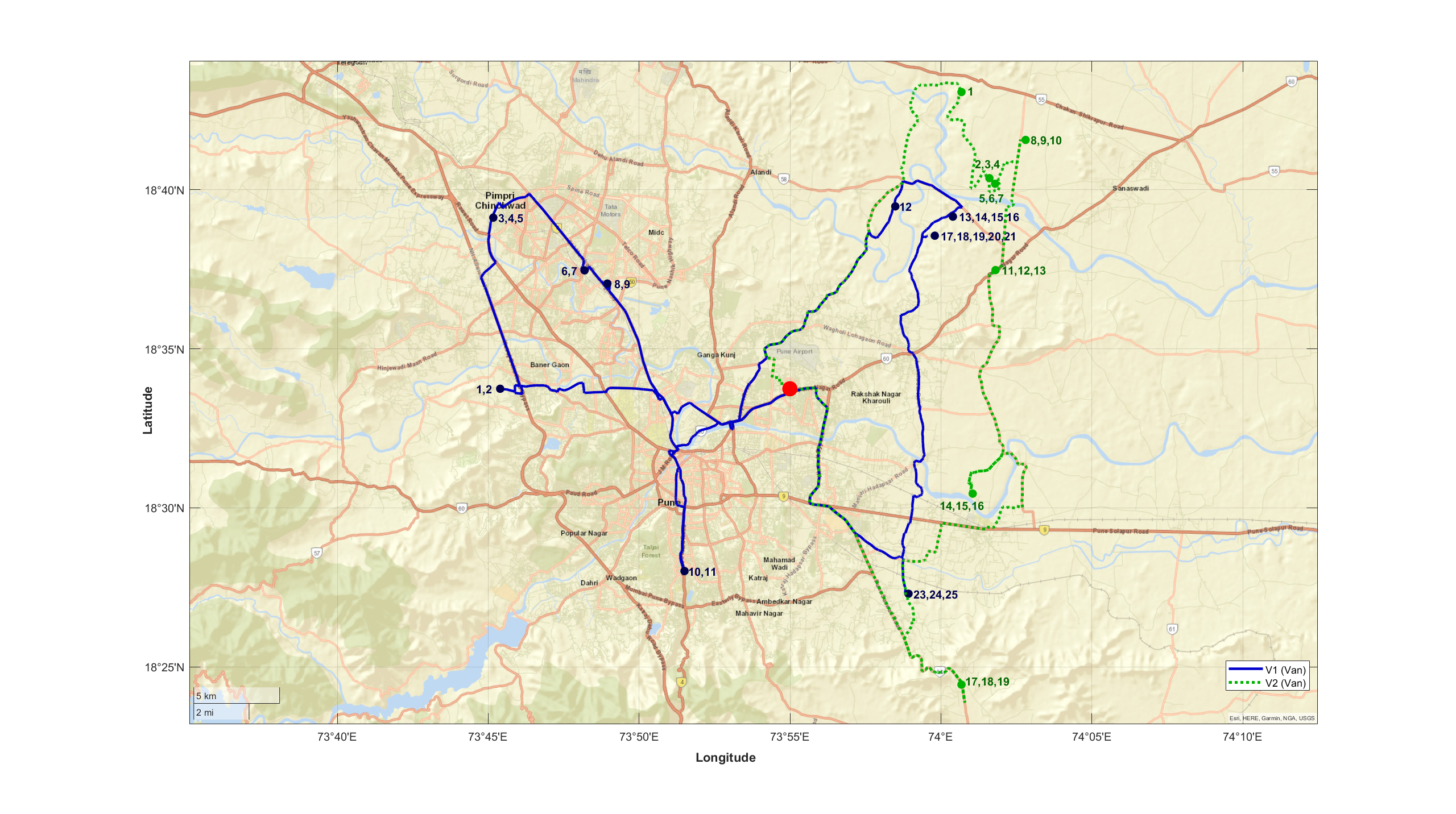}
    \caption{MRMH-HCVRP}
\end{subfigure}
\hfill
\begin{subfigure}[t]{0.49\textwidth}
    \centering
    \includegraphics[width=\linewidth,height=1.5\textheight,keepaspectratio]
    {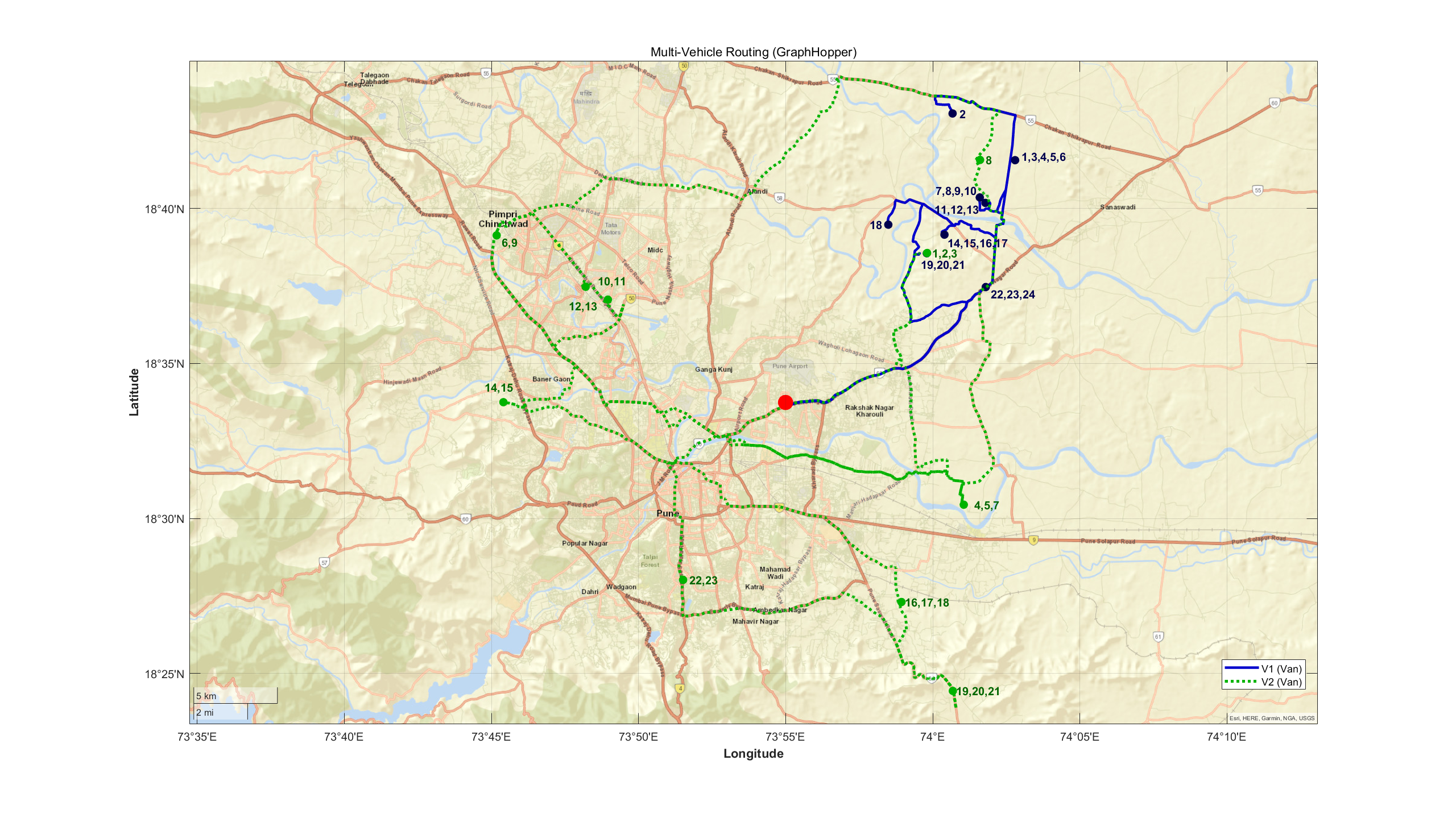}
    \caption{HCVRP}
\end{subfigure}

\caption{Comparison of four vehicle routing methods. All vehicles (vans) have identical capacities and serve the same set of customers. The routes of the two vehicles are represented by the blue and green lines, while the service sequence is indicated by the numbered labels at each customer location.}

\label{fig:4 method routes}

\end{sidewaysfigure}

Using order data with longer network-distances during normal periods (based on actual map-guided navigation routes rather than straight-line distances), we compared our proposed method against historical default sorting, shortest path (with same-location integration), and HCVRP (Figure \ref{fig:4 method routes}, Table \ref{tab:four_VRP_comparison}). The results demonstrate that our proposed approach successfully balances shorter routes and transportation times with maximized vehicle capacity utilization and the timely delivery of high-priority orders.


\begin{table}[h]
\centering
\footnotesize
\setlength{\tabcolsep}{5pt}
\renewcommand{\arraystretch}{1}
\caption{Comprehensive comparison of four vehicle routing methods}
\label{tab:four_VRP_comparison}
\begin{tabular}{lcccccc}
\hline
\multirow{2}{*}{Metric} & \multicolumn{4}{c}{Method} \\
\cmidrule(lr){2-5}
& \parbox{2.2cm}{\centering Time Sequence \\ \& Clustering} & \parbox{2.2cm}{\centering Shortest Path \\ \& Clustering} & HCVRP & \parbox{2.2cm}{\centering MRMH- \\ HCVRP} \\
\hline
\textbf{Vehicle 1} & & & & \\
\quad Distance (km) & 170.46 & 114.05 & 97.32 & \textbf{147.91} \\
\quad Time (hr) & 2.1307 & 1.4256 & 1.2165 & \textbf{1.8489} \\
\hline
\textbf{Vehicle 2} & & & & \\
\quad Distance (km) & 237.67 & 165.38 & 293.13 & \textbf{124.76} \\
\quad Time (hr) & 2.9709 & 2.0672 & 3.6642 & \textbf{1.5596} \\
\hline
\textbf{Total} & & & & \\
\quad Distance (km) & 408.13 & 279.43 & 390.45 & \textbf{272.67} \\
\quad Time (hr) & 5.1016 & 3.4928 & 4.8807 & \textbf{3.4085} \\
\hline
\textbf{High-Priority Orders} & & & & \\
\quad Rapid Delivery Rate (\%) & 85\% & 92\% & 38\% & \textbf{92\%} \\
\quad Priority First Ratio (\%) &89.36\% & 93.62\% & 89.36\% & \textbf{93.62\%} \\
\hline
\end{tabular}

\vspace{4pt} 
\scriptsize
\parbox{\linewidth}{
\textbf{Note:} 
\textit{Rapid Delivery Rate} means the highest priority or higher priority orders arrive ahead of their sensitive transit time without incurring any delay penalties and meeting customer expectations for fast delivery. 
\textit{Priority Delivery Ratio} means at the same point in time, based on the total quantity of all goods, the percentage of highest priority or higher priority orders that are scheduled for delivery earlier than other orders.
}
\end{table}

Among all the compared methods, the proposed MRMH-HCVRP achieved the most balanced and efficient performance. It ranks first in total distance (272.67 km) and total time (3.41 h), outperforming the Time Sequence \& Clustering method (408.13 km, 5.10 h) and HCVRP (390.45 km, 4.88 h), while also reducing the total distance slightly below the Shortest Path \& Clustering method (279.43 km). More importantly, the MRMH-HCVRP delivers the highest rapid delivery rate (92\%) and priority first ratio (93.62\%), tied with or exceeding the best competing methods. These results demonstrate that the proposed method not only improves routing efficiency but also effectively prioritizes high-priority orders, achieving a superior trade-off between operational cost and service quality.

\subsection{KPIs Improvement}
\subsubsection{Task Sequence and Allocation}
To illustrate the advantages of integrated optimization more clearly, we selected warehouse orders from three consecutive time periods. Owing to the priority of warehouse sorting, these three batches of orders were the lowest-ranked orders at 18:55, medium-ranked orders at 19:00, and best-ranked orders at 19:05. Owing to the limited throughput of the warehouse, each batch of orders experienced different average processing times within its respective time period: 15, 10, and 5 min. For out-of-warehouse shipping, these orders departed from the transportation origin (warehouse entrance) simultaneously and in the same batch. During the last-mile delivery process, reinforcement learning decisions are followed outside the warehouse to re-determine the order of picking and the route, ultimately delivering the goods to their destinations sequentially. Our model used one scooter (S1) and two motorcycles (M1 and M2) for road transport and performed delivery tasks during periods of moderate congestion. 

\begin{figure}[!ht]
\centering
  \includegraphics[width=0.9\textwidth]{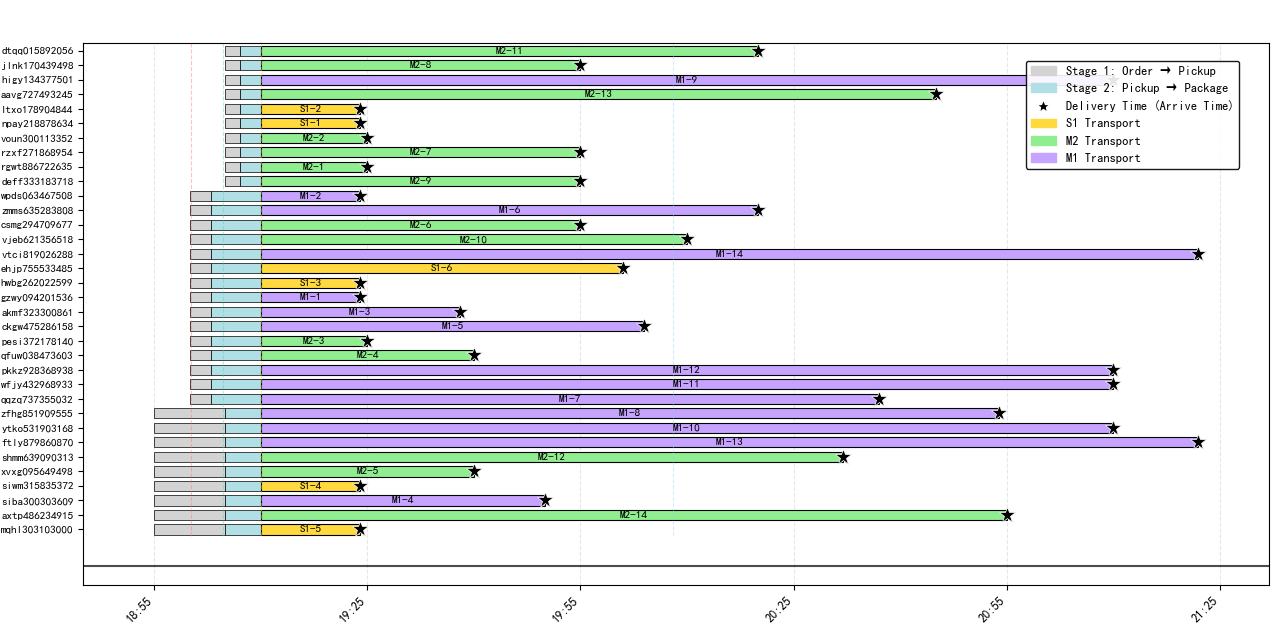}
  \caption{Gantt Chart for 34 orders}\label{fig:Gant}
\end{figure}

\begin{figure}[!ht]
\centering
  \includegraphics[width=0.8\textwidth]{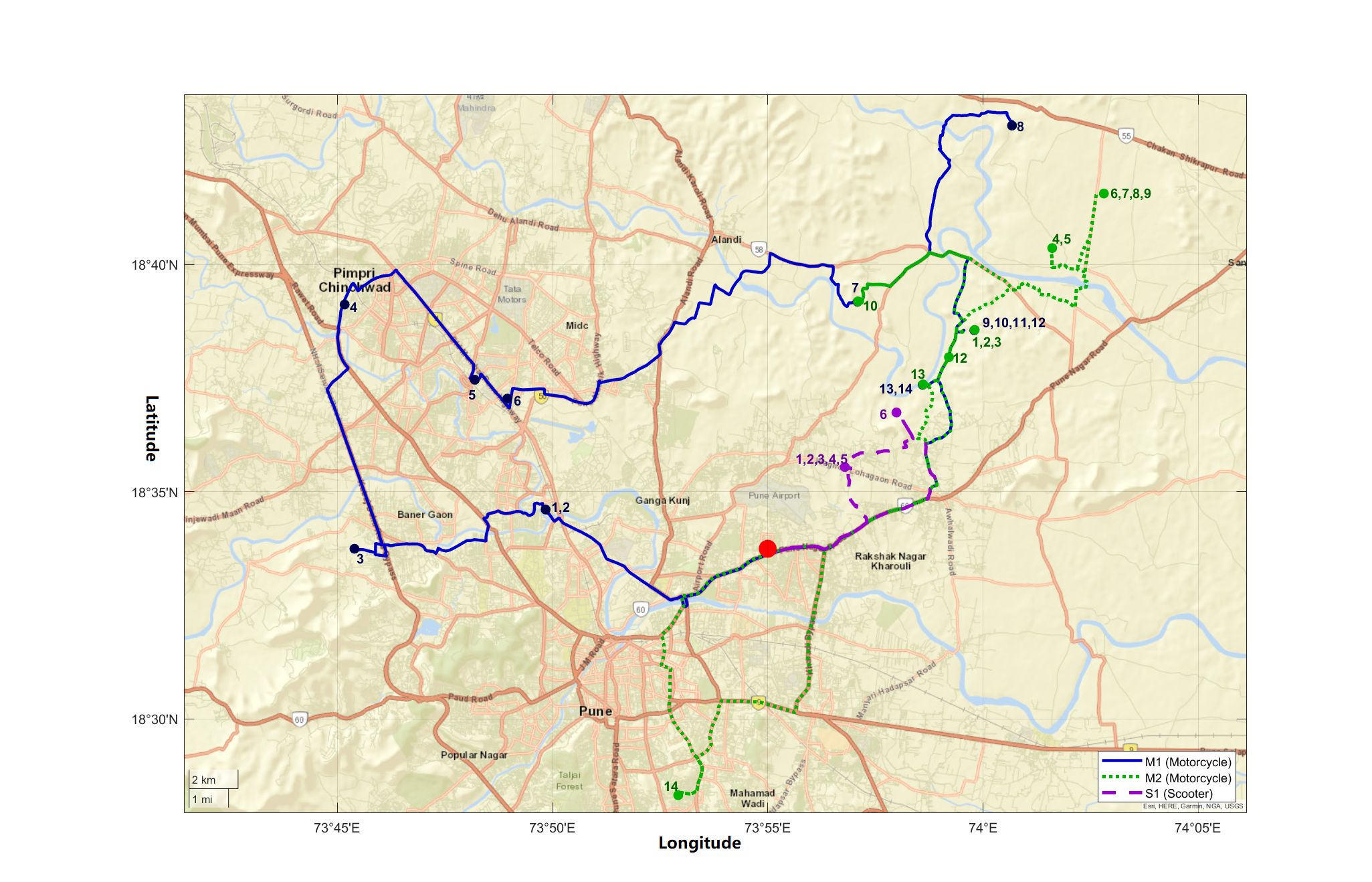}
  \caption{Routing for 34 orders. This route is generated from dynamic real-world map by Graphhooper.}
  
\label{fig:Comprehensive Route}
\end{figure}

The Gantt chart for the entire process is shown in Figure \ref{fig:Gant}, and the route allocation and specific connection sequence are shown in Figure \ref{fig:Comprehensive Route}. The proposed method significantly improves order grouping, pickup sequencing, and vehicle allocation, particularly for high-priority customers due to our targeted weight design. While baseline models perform adequately during low-volume periods when resources are easily allocated, our method demonstrates a pronounced advantage over baselines during peak hours by effectively managing traffic congestion, complex routing, and high order densities.

\subsubsection{Key KPIs Comparison}
For intelligent warehousing systems, key warehouse KPIs include, but are not limited to, order fulfillment time (the time from receiving an instruction to shipping), utilization rate of automated equipment (AGVs in this study), on-time inventory pickup rate,  and high-priority scheduling rate.
For last-mile delivery, key KPIs for land transportation include, but are not limited to, the average delivery time (the average time from leaving the warehouse to delivery to the customer), on-time delivery rate (the percentage of orders whose actual delivery time meets the promised time), high-priority satisfaction rate  (fast delivery for high-priority customers), vehicle utilization rate (full load condition), total travel distance, and latest delivery time.
For integration optimization, key KPIs, which are especially evident in the differences between independent optimization and traditional order services, consist of the following: high-priority customer service rate, average order completion time, longest completion time, and shortest completion time.

As shown in Table \ref{tab:KPI_comparison}, the proposed method consistently outperforms FIFO and historical baselines across all metrics. For inner warehouse operations, it achieves 100\% on-time departure and pickup rates alongside an 84.62\% high-priority scheduling rate, maintaining a competitive average fulfillment time of 9.85 minutes. For external last-mile transport, the method reduces average delivery time to 58 minutes and total distance to 209 km (a 46.4\% saving versus FIFO), while boosting the on-time delivery rate to 97.06\% and high-priority satisfaction to 92.31\%. Integrated system results further confirm these benefits, showing a 78.11\% comprehensive high-priority service rate and a 27.5\% reduction in average order completion time (66.38 minutes), demonstrating substantial system-wide improvements in efficiency, timeliness, and priority adherence.

\begin{table}[h]
\centering
\caption{Comparison of KPI Performance Under Different Optimization Strategies}
\label{tab:KPI_comparison}
\scriptsize
\begin{tabular}{p{1.8cm} p{3.5cm} p{1cm} p{1.5cm} p{1.5cm} p{1.2cm} p{2.2cm}}
\toprule
\textbf{Perspective} & \textbf{KPIs} & 
\textbf{Dir.} & 
\textbf{Proposed} & 
\textbf{FIFO} & 
\textbf{Historical Data} \\
\midrule

\multirow{4}{*}{In-Warehouse}
& Ave fulfilment time(mins) & $\downarrow$ & 9.85 & 9.68 & 10.74 \\
& On-time departure rate & $\uparrow$ & 100\% & 70.59\%  & 58.80\% \\
& High-priority scheduling rate & $\uparrow$ & 84.62\% & 76.92\% & 69.23\% \\
& On-time inventory pickup rate & $\uparrow$ & 100\% & 70.59\% & 73.53\% \\
\midrule

\multirow{6}{*}{Land Transport}
& Average delivery time(mins) & $\downarrow$ & 58 &  82  & 124 \\
& On-time delivery rate & $\uparrow$ & 97.06\% & 79.41\% & 76.47\%  \\
& High-priority satisfaction rate & $\uparrow$ & 92.31\% &  61.54\%  & 53.85\% \\
& Vehicle utilization rate & $\uparrow$ & 85.00\% & 75.50\%& --\\
& Total travel distance(km) & $\downarrow$ & 209 & 390 & -- \\
& Latest delivery time(mins) & $\downarrow$ & 132  & 217 & 265 \\
\midrule

\multirow{4}{*}{Joint Perspective}
& High-priority service rate & $\uparrow$ & 78.11\% & 46.95\% & 44.96\% \\
& Average order completion time & $\downarrow$ & 66.38 & 91.59 & -- \\
& Longest completion time  & $\downarrow$ & 142 & 227 & -- \\
& Shortest completion time  & $\downarrow$ & 19 & 26 & -- \\
\bottomrule
\end{tabular}
\end{table}

\section{Conclusions}\label{conclusions}
This study proposes an integrated multi-objective optimization framework that jointly addresses intelligent warehouse scheduling and last-mile delivery problems, aiming to balance efficiency, service quality, and operational robustness in complex logistics systems. By embedding multi-objective reinforcement learning into a hierarchical decision structure and combining it with heterogeneous vehicle routing planning, this framework surpasses traditional decompositional or single-objective methods, which fail to synchronize information from the internal and external environments and capture cross-stage influence relationships.

From a deep reinforcement learning methodology perspective, the proposed MORM-AGDQN introduces a structured multi-objective reward-shaping mechanism that stabilizes learning dynamics while maintaining sensitivity to competing operational objectives. Empirical results show that compared with similar reinforcement learning algorithms, this method achieves smoother convergence, higher cumulative rewards, and better scalability, particularly in environments with high dynamic demands and multiple agents working collaboratively. These findings demonstrate that multi-objective dynamic trade-offs during the learning process are crucial for achieving robust and stable decision-making strategies in stochastic logistics environments.

At the routing planning level, the proposed MRMH-HCVRP framework integrates distance- and traffic-condition-aware heterogeneous fleet allocation with a priority-sensitive service mechanism. The results show that this integrated design reduces the total travel distance and delivery time and significantly improves service differentiation for high-priority orders. Compared to traditional route planning strategies, this method achieves a more balanced trade-off between operating costs and service responsiveness (customer satisfaction), highlighting the importance of matching the vehicle performance with the space and demand characteristics.

From a holistic system perspective, KPI analysis confirms that the proposed integrated optimization strategy outperforms both FIFO  and historical strategies in both the warehousing and transportation phases. Notably, this framework achieves near-perfect just-in-time warehousing operations, significantly reduces delivery times, and substantially improves the service rate of high priority orders. These results strongly demonstrate that cross-decision-level coordination, which incorporates information from both inside and outside the warehouse, is crucial for improving overall system performance, particularly in high-density and time-sensitive logistics scenarios.

From a logistics management perspective, these findings provide actionable insights for practitioners. First, prioritizing high-value or time-sensitive orders within a multi-objective framework can significantly improve customer satisfaction without excessively increasing the operating costs. Second, heterogeneous fleet deployment based on delivery range and capacity constraints can improve dispatch flexibility, vehicle utilization efficiency, and energy efficiency. Third, integrating upstream and downstream decision-making processes and synchronizing dynamic core information enable more adaptive and resilient logistics operations under conditions of demand uncertainty and congestion.

Despite these contributions, some limitations exist. Current reinforcement learning frameworks rely on regional traffic information and predefined vehicle characteristics (based on historical data), which may not adequately reflect the volatility of the real-time operations. Future research can extend this work by incorporating segmented real-time traffic information, stochastic demand modelling, and adaptive fleet control strategies. Furthermore, the integration of emerging paradigms, such as digital twins or advanced learning architectures, can further enhance scalability and flexibility.
Given the NP-hard nature of this problem, exact optimization is computationally prohibitive for large-scale scenarios. While our RL framework provides efficient approximate solutions, future integration with quantum computing holds the potential to further improve solution quality and scalability beyond classical limits.


\section*{Acknowledgments}
This research was funded by a grant from the Canada Research Chair program in Disruptive Transportation Technologies and Services (CRC-2021-00480) and NSERC Discovery (RGPIN-2020-04492) fund.

\setlength{\bibsep}{0pt}
\bibliographystyle{elsarticle-harv}
\bibliography{traj.bib}

\newpage
\appendix
\section*{Appendix A: Pseudocode of Algorithms}
\label{app:pseudocode}
\addcontentsline{toc}{section}{Appendix A: Pseudocode of Algorithms}

\subsection*{A.1 Algorithm for Warehouse Operation}

\begin{algorithm}[!h]
\tiny
\caption{Framework of MORM-MADQN Algorithm}
\label{alg:cnn_main}
\begin{algorithmic}[1]
\State \textbf{Initialize:} 
\State \quad Q-networks $Q^i(\theta^i)$, target networks $Q^i({\theta^i}^{-})$ for all agents $i \in \mathcal{N}$
\State \quad CNN feature extractor $\phi$, CNN-Guider $\psi$, replay buffer $\mathcal{D}$
\State \quad Reward Machines: $\mathcal{RM}_{En}, \mathcal{RM}_{In}, \mathcal{RM}_{Ex}, \mathcal{RM}_{Sa}$

\For{episode = 1 to $M$}
    \State Reset environment, get initial state $\mathcal{S}_0$
    \State Extract features: $\vec{s}_0 \gets \text{CNN}(\mathcal{S}_0, \mathcal{P}_{\text{ex}}; \phi)$
    \State Initialize RM states: $u_0^{En}, u_0^{In}, u_0^{Ex}, u_0^{Sa}$

    \For{$t = 0$ to $T_{\text{max}}$}
        \For{each agent $i \in \mathcal{N}$}
            \State $\vec{p}^i_{\text{guide}} \gets \text{CNN-Guider}(\mathcal{O}^i; \psi)$
            \State $\hat{a}^i_{\text{CNN}} \gets \text{sample}(\vec{p}^i_{\text{guide}})$
            \State $\hat{a}^{i, A*}_t \gets \text{RM-A*}(\text{pos}(S^i), \text{goal}, \mathcal{RM}, \vec{p}^i_{\text{guide}})$
            \State $Q^i_{\text{RM}}(\vec{s}_t, a^i) \gets Q^i(\vec{s}_t, a^i; \theta^i) + \gamma_{\text{RM}} \cdot R_{\text{RM}}(a^i, \vec{u}_t)$
            \State Select $a^i_t$ using $\epsilon$-greedy on $Q^i_{\text{RM}}$
        \EndFor
        
        \State Execute $\vec{a}_t$, observe $\mathcal{S}_{t+1}$, rewards $\vec{r}_t$
        \State Extract features: $\vec{s}_{t+1} \gets \text{CNN}(\mathcal{S}_{t+1}, \mathcal{P}_{\text{ext}}'; \phi)$
        \State Update RM states: $\vec{u}_{t+1} \gets \mathcal{RM}(\vec{u}_t, \vec{a}_t, \vec{s}_{t+1})$
        \State Store $(\vec{s}_t, \vec{a}_t, \vec{r}_t, \vec{s}_{t+1}, \vec{u}_t, \vec{u}_{t+1})$ in $\mathcal{D}$
    \EndFor
\EndFor
\end{algorithmic}
\end{algorithm}

\begin{algorithm}[!h]
\tiny
\caption{Training of MORM-MADQN Algorithm}
\label{alg:cnn_train}
\begin{algorithmic}[2]
\Procedure{Training}{}
    \If{enough samples in $\mathcal{D}$}
        \State Sample mini-batch $\mathcal{B} \sim \mathcal{D}$
        \For{each sample $j$ and agent $i$}
            \State $\vec{f}^i_{\text{spatial}} \gets \text{CNN}_{\text{spatial}}(\mathcal{S}^{(j)})$
            \State $\vec{f}^i_{\text{pareto}} \gets \text{CNN}_{\text{pareto}}(\mathcal{P}_{\text{ext}}^{(j)})$
            \State $\vec{f}^i_{\text{Integrate}} \gets [\vec{f}^i_{\text{spatial}} \oplus \vec{f}^i_{\text{pareto}}]$
            
            \State $R^{i}_{\text{RM}} \gets \sum_k w_k \cdot \mathcal{RM}_k(a^{i,(j)}, u^{i,(j)}_k)$
            \State $y^{i,(j)} \gets R^{i}_{\text{RM}} + \gamma \max_{a^{'}} {Q^i}(\vec{s}^{\prime,(j)}, a^{'}, {{\theta}^{i}}^{-})$
            \State $\mathcal{L}^{i}_{\text{Q}} \gets (y^{i,(j)} - Q^i(\vec{s}^{(j)}, a^{i,(j)}; \theta^i))^2$
        \EndFor
        
        \State Update $\theta^i, \phi, \psi$ using $\nabla \mathcal{L}_{\text{Q}}$
        \State Update target networks: ${\theta^i}^{-} \gets \tau \theta^i + (1-\tau) {\theta^i}^{-}$
    \EndIf
\EndProcedure
\end{algorithmic}
\end{algorithm}

\subsection*{A.2 Algorithm for Last-Mile Vehicle Routing}

\begin{algorithm}[h]
\tiny
\caption{Framework of MORL for HCVRP}
\label{alg:vrp_main}
\begin{algorithmic}[3]
\Procedure{VRP-Routing}{$X, s_t$} 
    \State $h_N \leftarrow \Call{Encode}{X}$ 
    \State $(v, c) \leftarrow \Call{Decode}{s_t, h_N}$ 
    \State \textbf{return} $(v, c)$
\EndProcedure

\Procedure{Encode}{$X$}
    \State $X_{\text{orders}} = [\text{sequence, time, location}]$
    \State $X_{\text{fleet}} = [\text{type, capacity, location, speed, state}]$
    \State $X_{\text{traffic}} = [\text{congestion, delays, routes}]$
    \State $h_1 \leftarrow [X_{\text{orders}} \oplus X_{\text{fleet}} \oplus X_{\text{traffic}}]$
    \For{$l = 1$ to $N$}
        \State $h_{l+1} \leftarrow \text{MultiHeadAttention}(h_l)$
    \EndFor
    \State \textbf{return} $h_N$
\EndProcedure

\Procedure{Decode}{$s_t, h_N$}

\State \textbf{Vehicle Selection:}
    \For{each vehicle $h$}
        \State $\text{feasible}_h \gets (\text{not full capacity}) \land (\text{on time}) \land(\text{distance}_h)$
        \State $p^{veh}_h \gets \text{softmax}(W \cdot [h_N, \text{traffic}, \text{feasible}_h])$
    \EndFor
    \State $h \gets \text{sample}(\text{mask}(p^{veh}, \text{feasible}))$

\State \textbf{Customer Selection:}
    \For{each position $u$}
        \State $\text{feasible}_u \gets (\text{demand}_u \leq \text{cap}_h) \land (\text{delivery time}_u \text{ feasible})$
        \State $score[u] \gets \text{Compatibility}(h_N, u, \text{traffic})$
        \State $score[u] \gets score[u] + \text{MASK\_IF}(\neg \text{feasible}_u, -\infty)$
    \EndFor
    \State $u \gets \text{sample}(\text{softmax}(score))$

    \State \textbf{return} $(h, u)$
\EndProcedure

\end{algorithmic}
\end{algorithm}

\begin{algorithm}[h]
\tiny
\caption{Training of MORL for HCVRP}
\label{alg:vrp_train}
\begin{algorithmic}[4]
\State \textbf{Initialize:} Policy network $\pi_\theta$, Value network $V_\phi$, replay buffer $\mathcal{D}$

\For{episode = 1 to $M$}
    \State Reset environment, get initial state $s_0$
    \State Initialize empty solution
    
    \For{$t = 0$ to $T_{\text{max}}$}
        \State \textbf{Action Selection:}
        \State $(h_t, u_t) \gets \pi_\theta(s_t)$ \Comment{Select vehicle and customer}
        \State Execute action, observe next state $s_{t+1}$
        
        \State \textbf{Reward Computation:}
        \State $r_t \gets w_1 \cdot r_{\text{travel time}} + w_2 \cdot r_{\text{travel cost}} + w_3 \cdot r_{\text{priority}}$
        \State Store transition $(s_t, (h_t, u_t), r_t, s_{t+1})$ in $\mathcal{D}$
        
        \State \textbf{Training:}
        \If{enough samples in $\mathcal{D}$}
            \State Sample batch $\mathcal{B} \sim \mathcal{D}$
            \For{each transition in $\mathcal{B}$}
                \State Compute TD target: $y \gets r + \gamma V_\phi(s')$
                \State $\mathcal{L}_{\text{value}} \gets (V_\phi(s) - y)^2$
                \State $\mathcal{L}_{\text{policy}} \gets -\log \pi_\theta(a|s) \cdot (y - V_\phi(s))$
            \EndFor
            \State Update $\theta, \phi$ using gradient descent
        \EndIf
        
        \If{solution complete} \textbf{break} \EndIf
    \EndFor
\EndFor

\end{algorithmic}
\end{algorithm}


\end{document}